\documentclass[10pt,journal,compsoc]{IEEEtran}

\ifCLASSOPTIONcompsoc
\usepackage[nocompress]{cite}
\else
\usepackage{cite}
\fi
\ifCLASSINFOpdf
\usepackage[pdftex]{graphicx}
\else
\fi
\usepackage{amsmath}
\usepackage{array}
\usepackage{centernot}
\usepackage{multirow}
\usepackage{enumitem}
\usepackage{comment}
\usepackage[dvipsnames]{xcolor}

\newcommand{\Input}{{\hspace*{\algorithmicindent} \textbf{input }}}

\usepackage{algorithm}
\usepackage{algpseudocode}

\usepackage[nolist]{acronym}
\begin{acronym}[DRAM]
    \acro{AI}{Artificial Intelligence}
    \acrodefindefinite{AI}{an}{an}
    \acro{ASIC}{application-specific integrated circuit}
    \acrodefindefinite{ASIC}{an}{an}
    \acro{BD}{bounded delay}
    \acro{BNN}{binarized neural network}
    \acro{CNN}{convolutional neural network}
    \acro{CTM}{convolutional Tsetlin machine}
    \acro{FSM}{finite state machine}
    \acro{LA}{learning automaton}
    \acrodefplural{LA}{learning automata}
    \acrodefindefinite{LA}{an}{a}
    \acro{ML}{machine learning}
    \acrodefindefinite{ML}{an}{a}
    \acro{TA}{Tsetlin automaton}

    \acrodefplural{TA}{Tsetlin automata}
    \acro{TAT}{Tsetlin automaton team}
    \acro{TM}{Tsetlin machine}
    \acro{RTM}{regression Tsetlin machine}
\end{acronym}

\begin{document}
\title{A Relational Tsetlin Machine with Applications to Natural Language Understanding}
\author{Rupsa~Saha, Ole-Christoffer~Granmo,
Vladimir~I.~Zadorozhny,
Morten~Goodwin
\IEEEcompsocitemizethanks{\IEEEcompsocthanksitem R.~Saha, O.~C.~Granmo and M.~Goodwin are with Centre for AI Research, Department of IKT, University of Agder, Norway.\protect\\
\IEEEcompsocthanksitem V.~I.~Zadorozhny is with School of Computing and Information, University of Pittsburgh, USA, and Centre for AI Research, University of Agder, Norway.}
}
\ifCLASSOPTIONpeerreview
\markboth{arXiv preprint}%
{A Relational \ac{TM} with Applications to Natural Language Understanding}
\else
\markboth{arXiv preprint}%
{Saha \MakeLowercase{\textit{et al.}}: A Relational \ac{TM} with Applications to Natural Language Understanding}
\fi

\IEEEtitleabstractindextext{%
\begin{abstract}
\acp{TM} are a pattern recognition approach that uses finite state machines for learning and propositional logic to represent patterns. In addition to being natively interpretable, they have provided competitive accuracy for various tasks. In this paper, we increase the computing power of \acp{TM} by proposing a first-order logic-based framework with Herbrand semantics. The resulting \ac{TM} is \emph{relational} and can take advantage of logical structures appearing in natural language, to learn rules that represent how actions and consequences are related in the real world. The outcome is a logic program of Horn clauses, bringing in a structured view of unstructured data. In closed-domain question-answering, the first-order representation produces $10\times$ more compact KBs, along with an increase in  answering accuracy from $94.83\%$ to $99.48\%$. The approach is further robust towards erroneous, missing, and superfluous information, distilling the aspects of a text that are important for real-world understanding.\end{abstract}

}
\maketitle

\IEEEdisplaynontitleabstractindextext

%
\IEEEpeerreviewmaketitle

\ifCLASSOPTIONcompsoc
\IEEEraisesectionheading{\section{Introduction}\label{sec:introduction}}
\else
\section{Introduction}
\label{sec:introduction}
\fi
\IEEEPARstart{U}{sing} \ac{AI} to answer natural language questions has long been an active research area, considered as an essential aspect in machines ultimately achieving human-level world understanding. Large-scale structured knowledge bases (KBs), such as Freebase \cite{bollacker2008freebase}, have been a driving force behind successes in this field. The KBs encompass massive ever-growing amounts of information, which enable easier handling of Open-Domain Question-Answering (QA) \cite{prager2006open} by organizing a large variety of answers in a structured format. The difficulty arises in successfully interpreting natural language by artificially intelligent agents, both to build the KBs from natural language text resources and to interpret the questions asked.

Generalization beyond the information stored in a KB further complicates the QA problem. Human-level world understanding requires abstracting from specific examples to build more general concepts and rules. When the information stored in the KB is error-free and consistent, generalization becomes a standard inductive reasoning problem. However, abstracting world-knowledge entails dealing with uncertainty, vagueness, exceptions, errors, and conflicting information. This is particularly the case when relying on \ac{AI} approaches to extract and structure information, which is notoriously error-prone.

This paper addresses the above QA challenges by proposing a Relational \ac{TM} that builds non-recursive first-order \emph{Horn clauses} from specific examples, distilling general concepts and rules.

\textbf{Tsetlin Machines} \cite{granmo2018tsetlin} are a pattern recognition approach to constructing human-understandable patterns from given data, founded on propositional logic. While the idea of \ac{TA} \cite{tsetlin1961behaviour} have been around since 1960s, using them in pattern recognition is relatively new. \acp{TM} have successfully addressed several machine learning tasks, including natural language understanding \cite{yadav2021sentiment,yadav2021wordsense,bhattarai2021novelty,saha2020causal,berge2019using}, image analysis \cite{granmo2019convolutional}, classification \cite{abeyrathna2021integer}, regression \cite{abeyrathna2019nonlinear}, and speech understanding \cite{lei2021kws}. 
The propositional clauses constructed by a \ac{TM} have high discriminative power and constitute a global description of the task learnt \cite{blakely2020closedform,saha2020causal}. Apart from maintaining accuracy comparable to state-of-the-art machine learning techniques, the method also has provided a smaller memory footprint and faster inference than more traditional neural network-based models~\cite{wheeldon2020pervasive,lei2020arithmetic,abeyrathna2021integer,lei2021kws}. Furthermore, \cite{shafik2020explainability} shows that \acp{TM} can be fault-tolerant, able to mask stuck-at faults. However, although \acp{TM} can express any propositional formula by using disjunctive normal form, first-order logic is required to obtain the computing power equivalent to a universal Turing machine. In this paper, we take the first steps towards increasing the computing power of \acp{TM} by introducing a \emph{first order} \ac{TM} framework with Herbrand semantics, referred to as the \emph{Relational} \ac{TM}. Accordingly, we will in the following denote the original approach as \emph{Propositional \acp{TM}}.

\textbf{Closed-Domain Question-Answering:} As proof-of-concept, we apply our proposed Relational \ac{TM} to so-called Closed-Domain QA. Closed-Domain QA assumes a text (single or multiple sentences) followed by a question which refers to some aspect of the preceding text. Accordingly, the amount of information that must be navigated is less than for open question-answering. Yet, answering closed-domain questions poses a significant natural language understanding challenge.

Consider the following example of information, taken from \cite{rajpurkar2016squad}: ``The Black Death is thought to have originated in the arid plains of Central Asia, where it then travelled along the Silk Road, reaching Crimea by 1343. From there, it was most likely carried by Oriental rat fleas living on the black rats that were regular passengers on merchant ships."One can then have questions such as ``Where did the black death originate?" or ``How did the black death make it to the Mediterranean and Europe?". These questions can be answered completely with just the information provided, hence it is an example of closed-domain question answering. However, mapping the question to the answer requires not only natural language processing, but also a fair bit of language understanding.

Here is a much simpler example: ``Bob went to the garden. Sue went to the cafe. Bob walked to the office." This information forms the basis for questions like ``Where is Bob?" or ``Where is Sue?". Taking it a step further, given the previous information and questions, one can envision a model that learns to answer similar questions based on similar information, even though the model has never seen the specifics of the information before (i.e., the names and the locations).

With QA being such an essential area of Natural Language Understanding, there has been a lot of different approaches proposed. Common methods to QA include the following:
\begin{itemize}
\item Linguistic techniques, such as tokenization, POS tagging and parsing
that transform questions into a precise query that merely extracts the respective response from a structured database;
\item Statistical techniques such as Support Vector Machines, Bayesian Classifiers, and maximum entropy models, trained on large amount of data, specially for open QA;
\item Pattern matching using surface text patterns with templates for response generation.
\end{itemize}
Many methods use a hybrid approach encompassing more than one of these approaches for increased accuracy. Most QA systems suffer from a lack of generality, and are tuned for performance in restricted use cases. Lack of available explainabilty also hinders researchers' quest to identify pain points and possible major improvements \cite{soares2020literature, pundge2016question}.

\textbf{Paper Contributions:}
Our main contributions in this paper are as follows:
\begin{itemize}
\item We introduce a \emph{Relational} \ac{TM}, as opposed to a propositional one, founded on non-recursive Horn clauses and capable of processing relations, variables and constants.
\item We propose an accompanying relational framework for efficient representation and processing of the QA problem.
\item We provide empirical evidence uncovering that the Relational \ac{TM} produces at least one order of magnitude more compact KBs than the Propositional \ac{TM}. At the same time, answering accuracy increases from $94.83$\% to $99.48$\% because of more general rules.
\item We provide a model-theoretical interpretation for the proposed framework.
\end{itemize}
Overall, our Relational \ac{TM} unifies knowledge representation, learning, and reasoning in a single framework.

\textbf{Paper Organization:}
The paper is organized as follows. In Section \ref{sec:related}, we present related work on Question Answering. Section \ref{sec:reltm} focuses on the background of the Propositional \ac{TM} and the details of the new Relational \ac{TM}. In Sections \ref{sec:qa in relTM} and \ref{sec:experiments}, we describe how we employ Relational \acp{TM} in QA and related experiments.

\section{Background and Related Work}\label{sec:related}
The problem of QA is related to numerous aspects of Knowledge Engineering and Data Management.

Knowledge engineering deals with constructing and maintaining knowledge bases to store knowledge of the real world in various domains. Automated reasoning techniques use this knowledge to solve problems in domains that ordinarily require human logical reasoning. Therefore, the two key issues in knowledge engineering are how to construct and maintain knowledge bases, and how to derive new knowledge from existing knowledge effectively and efficiently. Automated reasoning is concerned with the building of computing systems that automate this process. Although the overall goal is to automate different forms of reasoning, the term has largely been identified with valid deductive reasoning as conducted in logical systems. This is done by combining known (yet possibly incomplete) information with background knowledge and making inferences regarding unknown or uncertain information. 

Typically, such a system consists of subsystems like knowledge acquisition system, the knowledge base itself, inference engine, explanation subsystem and user interface. The knowledge model has to represent the relations between multiple components in a symbolic, machine understandable form, and the inference engine has to manipulate those symbols to be capable of reasoning. The ``way to reason" can range from earlier versions that were simple rule-based systems to more complex and recent approaches based on machine learning, especially on Deep Learning. Typically, rule-based systems suffered from lack of generality, and the need for human experts to create rules in the first place. On the other hand most machine learning based approaches have the disadvantage of not being able to justify decisions taken by them in human understandable form \cite{ludwig2010comparison, cyras2020machine}.

While databases have long been a mechanism of choice for storing information, they only had inbuilt capability to identify relations between various components, and did not have the ability to support reasoning based on such relations. Efforts to combine formal logic programming with relational databases led to the advent of deductive databases. In fact, the field of QA is said to have arisen from the initial goal of performing deductive reasoning on a set of given facts~\cite{green1969theorem}. In deductive databases, the semantics of the information are represented in terms of mathematical logic. Queries to deductive databases also follow the same logical formulation \cite{gallaire1989logic}. One such example is ConceptBase \cite{jarke1995conceptbase}, which used the Prolog-inspired language O-Telos for logical knowledge representation and querying using deductive object-oriented database framework.

With the rise of the internet, there came a need for unification of information on the web. The Semantic Web (SW) proposed by W3C is one of the approaches that bridges the gap between the Knowledge Representation and the Web Technology communities. However, reasoning and consistency checking is still not very well developed, despite the underlying formalism that accompanies the semantic web. One way of introducing reasoning is via descriptive logic. It involves concepts (unary predicates) and roles (binary predicates) and the idea is that implicitly captured knowledge can be inferred from the given descriptions of concepts and roles \cite{dong2003checking, turhan2011description}.

One of the major learning exercises is carried out by the NELL mechanism proposed by \cite{mitchell2018never}, which aims to learn many semantic categories from primarily unlabeled data. At present, NELL uses simple frame-based knowledge representation, augmented by the PRA reasoning system. The reasoning system performs tractable, but limited types of reasoning based on restricted Horn clauses. NELL’s capabilities is already limited in part by its lack of more powerful reasoning components; for example, it currently lacks methods for representing and reasoning about time and space. Hence, core \ac{AI} problems of representation and tractable reasoning are also core research problems for never-ending learning agents.

While other approaches such as neural networks are considered to provide attribute-based learning, Inductive Logic Programming (ILP) is an attempt to overcome their limitations by moving the learning away from the attributes themselves and more towards the level of first-order predicate logic. ILP builds upon the theoretical framework of logic programming and looks to construct a predicate logic given background knowledge, positive examples and negative examples. One of the main advantages of ILP over attribute-based learning is ILP’s generality of representation for background knowledge. This enables the user to provide, in a more natural way, domain-specific background knowledge to be used in learning. The use of background knowledge enables the user both to develop a suitable problem representation and to introduce problem-specific constraints into the learning process. 
Over the years, ILP has evolved from depending on hand-crafted background knowledge only, to employing different technologies in order to learn the background knowledge as part of the process. In contrast to typical machine learning, which uses feature vectors, ILP requires the knowledge to be in terms of facts and rules governing those facts. Predicates can either be supplied or deducted, and one of the advantages of this method is that newer information can be added easily, while previously learnt information can be maintained as required \cite{cropper2020turning}.
Probabilistic inductive logic programming is an extension of ILP, where logic rules, as learnt from the data, are further enhanced by learning probabilities associated with such rules \cite{bratko1995applications, nickles2014probabilistic, de2008probabilistic}. 

To sum up, none of the above approaches can be efficiently and systematically applied to the QA problem, especially in uncertain and noisy environments. In this paper we propose a novel approach to tackle this problem. Our approach is based on relational representation of QA, and using a novel Relational \ac{TM} technique for answering questions. We elaborate on the proposed method in the next two sections.

\section{Building a Relational Tsetlin Machine}
\label{sec:reltm}
\subsection{Tsetlin Machine Foundation}\label{sec:TM_foundation}

A Tsetlin Automaton (\ac{TA}) is a deterministic automaton that learns the optimal action among the set of actions offered by an environment. It performs the action associated with its current state, which triggers a reward or penalty based on the ground truth. The state is updated accordingly, so that the \ac{TA} progressively shifts focus towards the optimal action~\cite{tsetlin1961behaviour}. A \ac{TM} consists of a collection of such TAs, which together create complex propositional formulas using conjunctive clauses.

\subsubsection{Classification}

A \ac{TM} takes a vector $X=(x_1,\ldots,x_f)$ of propositional variables as input, to be classified into one of two classes, $y=0$ or $y=1$. Together with their negated counterparts, $\bar{x}_k = \lnot x_k = 1-x_k$, the features form a literal set $L = \{x_1,\ldots,x_f,\bar{x}_1,\ldots,\bar{x}_f\}$. We refer to this ``regular'' TM as a Propositional TM, due to the input it works with and the output it produces.

A \ac{TM} pattern is formulated as a conjunctive clause $C_j$, formed by ANDing a subset $L_j \subseteq L$ of the literal set:
\begin{equation}
\textstyle
C_j (X)=\bigwedge_{l_k \in L_j} l_k = \prod_{l_k \in L_j} l_k.
\end{equation}
E.g., the clause $C_j(X) = x_1 \land x_2 = x_1 x_2$ consists of the literals $L_j = \{x_1, x_2\}$ and outputs $1$ iff $x_1 = x_2 = 1$.

The number of clauses employed is a user set parameter~$n$. Half of the $n$ clauses are assigned positive polarity ($C_j^+$). The other half is assigned negative polarity ($C_j^-$). The clause outputs, in turn, are combined into a classification decision through summation:
\begin{equation}
\textstyle
v = \sum_{j=1}^{n/2} C_j^+(X) - \sum_{j=1}^{n/2} C_j^-(X).
\end{equation}
In effect, the positive clauses vote for $y=1$ and the negative for $y=0$. Classification is performed based on a majority vote, using the unit step function: $\hat{y} = u(v) = 1 ~\mathbf{if}~ v \ge 0 ~\mathbf{else}~ 0$. The classifier
$\hat{y} = u\left(x_1 \bar{x}_2 + \bar{x}_1 x_2 - x_1 x_2 - \bar{x}_1 \bar{x}_2\right)$, for instance, captures the XOR-relation.

\subsubsection{Learning}

Alg. \ref{algo:tm} encompasses the entire learning procedure. We observe that, learning is performed by a team of $2f$ \acp{TA} per clause, one \ac{TA} per literal $l_k$ (Alg. \ref{algo:tm}, Step~\ref{initialization}). Each \ac{TA} has two actions -- Include or Exclude -- and decides whether to include its designated literal $l_k$ in its clause. 

\acp{TM} learn on-line, processing one training example $(X, y)$ at a time (Step~\ref{inputstep}). The \acp{TA} first produce a new configuration of clauses (Step~\ref{includeexcludestep}), $C_1^+, \ldots, C_{n/2}^-$, followed by calculating a voting sum $v$ (Step~\ref{predictstep}).

Feedback are then handed out stochastically to each \ac{TA} team.  The difference $\epsilon$ between the clipped voting sum $v^c$ and a user-set voting target $T$ decides the probability of each \ac{TA} team receiving feedback (Steps~\ref{feedbackstart}-\ref{feedbackstop}). Note that the voting sum is clipped to normalize the feedback probability. The voting target for $y=1$ is $T$ and for $y=0$ it is $-T$. Observe that for any input $X$, the probability of reinforcing a clause gradually drops to zero as the voting sum approaches the user-set target. This ensures that clauses distribute themselves across the frequent patterns, rather than missing some and over-concentrating on others. 

Clauses receive two types of feedback. Type I feedback produces frequent patterns, while Type II feedback increases the discrimination power of the patterns.

\textbf{Type I feedback} is given stochastically to clauses with positive polarity when $y=1$ and to clauses with negative polarity when $y=0$. Each clause, in turn, reinforces its \acp{TA} based on: (1) its output $C_j(X)$; (2) the action of the \ac{TA} -- Include or Exclude; and (3) the value of the literal $l_k$ assigned to the \ac{TA}. Two rules govern Type I feedback:
\begin{itemize}
\item \emph{Include} is rewarded and \emph{Exclude} is penalized with probability $\frac{s-1}{s}$ whenever $C_j(X)=1~\mathbf{and}~l_k=1$. This reinforcement is strong (triggered with high probability) and makes the clause remember and refine the pattern it recognizes in $X$.\footnote{Note that the probability $\frac{s-1}{s}$ is replaced by $1$ when boosting true positives.} 
\item \emph{Include} is penalized and \emph{Exclude} is rewarded with probability $\frac{1}{s}$ whenever $C_j(X)=0~\mathbf{or}~l_k=0$. This reinforcement is weak (triggered with low probability) and coarsens infrequent patterns, making them frequent.
\end{itemize}
Above, the user-configurable parameter $s$ controls pattern frequency, i.e., a higher $s$ produces less frequent patterns.

\textbf{Type II feedback} is given stochastically to clauses with positive polarity when $y=0$ and to clauses with negative polarity when $y\!=\!1$. It penalizes \emph{Exclude} whenever $C_j(X)=1~\mathbf{and}~l_k=0$. Thus, this feedback produces literals for discriminating between $y\!=\!0$ and $y=1$, by making the clause evaluate to $0$ when facing its competing class. Further details can be found in \cite{granmo2018tsetlin}.



\begin{algorithm}[t]
\small
\caption{Propositional \ac{TM}}
\label{algo:tm}

\Input{Tsetlin Machine $\mathrm{TM}$, Example pool $S$, Training rounds $e$, Clauses $n$, Features $f$, Voting target $T$, Specificity $s$}

\begin{algorithmic} [1]
\Procedure{Train}{$\mathrm{TM}, S, e, n, f, T, s$}
\For{$j \gets 1, \ldots, n/2$}\label{initialization}
\State $\mathit{TA}_j^+ \gets \mathrm{RandomlyInitializeClauseTATeam}(2f)$
\State $\mathit{TA}_j^- \gets \mathrm{RandomlyInitializeClauseTATeam}(2f)$
\EndFor
\For{$i \gets 1, \ldots, e$}
\State $(X_i, y_i) \gets \mathrm{ObtainTrainingExample}(S)$\label{inputstep}
\State $C_1^+, \ldots, C_{n/2}^- \gets \mathrm{ComposeClauses}(\mathit{TA}_1^+,\ldots,\mathit{TA}_{n/2}^-)$\label{includeexcludestep}
\State $v_i \gets \sum_{j=1}^{n/2} C_j^+(X_i) - \sum_{j=1}^{n/2} C_j^-(X_i)$ \Comment{Vote sum}\label{predictstep}
\State $v_i^c \leftarrow \mathbf{clip}\left(v_i, -T, T\right)$ \Comment{Clipped vote sum}

\For{$j \gets 1, \ldots, n/2$} \Comment{Update \ac{TA} teams}
\If{$y_i = 1$}\label{feedbackstart}
    \State $\epsilon \gets T - v_i^c$ \Comment{Voting error}
    \State TypeIFeedback($X_i, \mathit{TA}_j^+, s$) \textbf{if} rand() $\le \frac{\epsilon}{2T}$
    \State TypeIIFeedback($X_i, \mathit{TA}_j^-$) \textbf{if} rand() $\le \frac{\epsilon}{2T}$
\Else
    \State $\epsilon \gets T + v_i^c$ \Comment{Voting error}
    \State TypeIIFeedback($X_i, \mathit{TA}_j^+$) \textbf{if} rand() $\le \frac{\epsilon}{2T}$
    \State TypeIFeedback($X_i, \mathit{TA}_j^-, s$) \textbf{if} rand() $\le \frac{\epsilon}{2T}$
\EndIf\label{feedbackstop}
\EndFor
\EndFor
\EndProcedure
\end{algorithmic}
\end{algorithm}

\begin{figure*}[ht]
\centering
\includegraphics[width=0.9\linewidth]{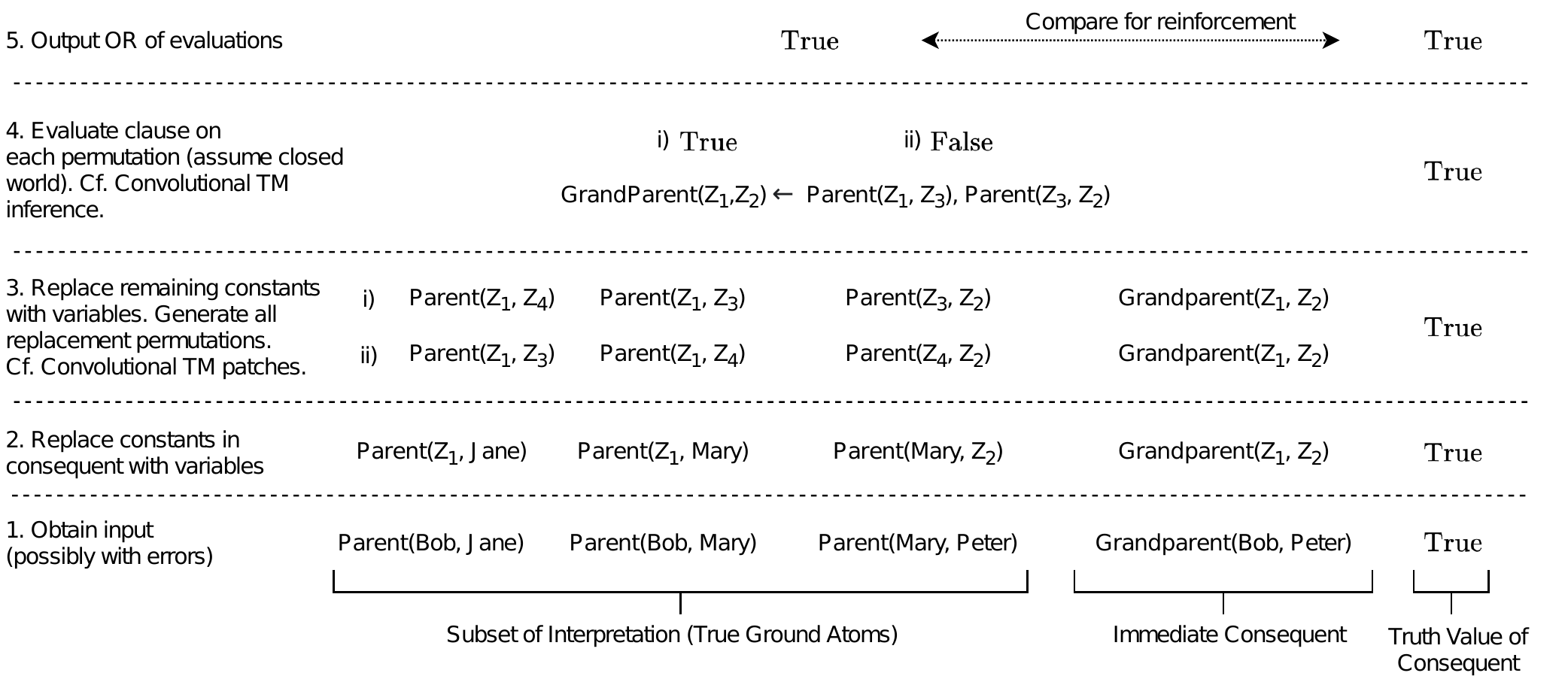}
\caption{Relational \ac{TM} processing steps}
\label{figure:tm_steps}
\end{figure*}

\begin{algorithm}[t]
\small
\caption{Relational \ac{TM}}
\label{algo:rtm}

\Input{Convolutional Tsetlin Machine $\mathrm{TM}$, Example pool $S$, Number of training rounds $e$}

\begin{algorithmic} [1]
\Procedure{Train}{$\mathrm{TM}, S, e$}

\For{$i \gets 1, \ldots, e$}
\State $(\tilde{\mathcal{X}}, \tilde{\mathcal{Y}}) \gets \mathrm{ObtainTrainingExample}(S)$
\State $A' \gets \mathrm{ObtainConstants}(\tilde{\mathcal{Y}})$
\State $(\tilde{\mathcal{X}}',\tilde{\mathcal{Y}}') \gets \mathrm{VariablesReplaceConstants}(\tilde{\mathcal{X}}, \tilde{\mathcal{Y}}, A')$
\State $A'' \gets \mathrm{ObtainConstants}(\tilde{\mathcal{X}}')$
\State $Q \gets \mathrm{GenerateVariablePermutations}(\tilde{\mathcal{X}}', A'')$
\State $\mathrm{UpdateConvolutionalTM}(\mathrm{TM}, Q, \tilde{\mathcal{Y}}')$
\EndFor

\EndProcedure
\end{algorithmic}
\end{algorithm}

\subsection{Relational Tsetlin Machine}

In this section, we introduce the Relational \ac{TM}, which a major contribution of this paper. It is designed to take advantage of logical structures appearing in natural language and process them in a way that leads to a compact, logic-based representation, which can ultimately reduce the gap between structured and unstructured data. While the Propositional \ac{TM} operates on propositional input variables $X = (x_1 \ldots, x_f)$, building propositional conjunctive clauses, the Relational \ac{TM} processes relations, variables and constants, building Horn clauses. Based on Fig.~\ref{figure:tm_steps} and Alg.~\ref{algo:rtm}, we here describe how the Relational \ac{TM} 
builds upon the original \ac{TM} in three steps. 
First, we establish an approach for dealing with relations and constants. This is done by mapping the relations to propositional inputs, allowing the use of a standard \ac{TM}. We then introduce Horn clauses with variables, showing how this representation detaches the \ac{TM} from the constants, allowing for a more compact representation compared to only using propositional clauses. We finally introduce a novel convolution scheme that effectively manages multiple possible mappings from constants to variables. While the mechanism of convolution remains the same as in the original \cite{granmo2019convolutional}, what we wish to attain from using it in a Relational \ac{TM} context is completely different, as explained in the following.

\subsubsection{Model-theoretical Interpretation}

The concept of Relational \ac{TM} can be grounded in the model-theoretical interpretation of a logic program without functional symbols and with a finite {\em Herbrand model} \cite{lloyd1984, KOWALSKI2014523}. The ability to represent learning in the form of Horn clauses is extremely useful due to the fact that Horn clauses are both simple, as well as powerful enough to describe any logical formula \cite{KOWALSKI2014523}.

Next, we define the {\em Herbrand model} of a logic program. A {\em Herbrand Base (HB)} is the set of all possible ground atoms, i.e., atomic formulas without variables, obtained using predicate names and constants in a logic program {\em P}.
A {\em Herbrand Interpretation} is a subset $I$ of the {\em Herbrand Base} ($I \subseteq HB$). 
To introduce the {\em Least Herbrand Model} we define the immediate consequence operator $TP: P(HB) \rightarrow P(HB)$, which for an {\em Herbrand Interpretation} $I$ produces the interpretation that immediately follows from $I$ by the rules (Horn clauses) in the program $P$:
\begin{eqnarray*}
TP(I)&= \{A_0 \in HB \mid A_0 \gets A_1,...,A_n\nonumber\\
&\in ground(P) \wedge \{A_1,...,A_n\} \subseteq I\} \cup I.
\end{eqnarray*}

The least fixed point {\em lfp(TP)} of the immediate consequence operator with respect to subset-inclusion is the {\em Least Herbrand Model (LHM)} of the program $P$. {\em LHM} identifies the semantics of the program $P$: it is the {\em Herbrand Interpretation} that contains those and only those atoms that follow from the program:
\begin{equation*}
\forall A \in HB: P \models A \Leftrightarrow A \in \mathrm{LHM}.
\end{equation*}

As an example, consider the following program $P$:
\begin{eqnarray*}
p(a).~q(c).\\
q(X) \leftarrow p(X).
\end{eqnarray*}
Its Herbrand base is
\begin{equation*}
HB = \{p(a), p(c), q(a), q(c)\},
\end{equation*}
and its Least Herbrand Model is:
\begin{equation*}
 \mathrm{LHM} = \mathrm{lfp}(TP) =\{p(a), q(a), q(c)\},
\end{equation*}
which is the set of atoms that follow from the program $P$.

\subsubsection{Learning Problem}

Let $A = \{a_1, a_2, \ldots, a_q\}$ be a finite set of constants and let $R = \{r_1, r_2, \ldots, r_p\}$ be a finite set of relations of arity $w_u \ge 1, u \in \{1, 2, \ldots, p\}$, which forms the alphabet $\Sigma$. The Herbrand base\begin{eqnarray}
HB &= \{r_1(a_1, a_2, \ldots,a_{w_1}), r_1(a_2, a_1, \ldots,a_{w_1}),\nonumber\\
&\quad\ldots, r_p(a_1, a_2, \ldots,a_{w_p}), r_p(a_2, a_1, \ldots,a_{w_p}), \ldots\}
\end{eqnarray}
is then also finite, consisting of all the ground atoms that can be expressed using $A$ and $R$.

We also have a logic program $P$, with program rules expressed as Horn clauses without recursion. Each Horn clause has the form:
\begin{equation}
B_0 \leftarrow B_1, B_2, \cdots, B_d.
\end{equation}
Here, $B_l, l \in \{0, \ldots, d\},$ is an atom $r_u(Z_1, Z_2, \ldots,Z_{w_u})$ with variables $Z_1, Z_2, \ldots,Z_{w_u}$, or its negation $\lnot r_u(Z_1, Z_2, \ldots,Z_{w_u})$. The arity of $r_u$ is denoted by $w_u$.

Now, let $\mathcal{X}$ be a subset of the LHM of $P$, $\mathcal{X} \subseteq \mathrm{lfp}(TP)$, and let $\mathcal{Y}$ be the subset of the LHM that follows from $\mathcal{X}$ due to the Horn clauses in $P$. Further assume that atoms are randomly removed and added to $\mathcal{X}$ and $\mathcal{Y}$ to produce a possibly noisy observation 
$(\tilde{\mathcal{X}}, \tilde{\mathcal{Y}})$, i.e., $\tilde{\mathcal{X}}$ and $\tilde{\mathcal{Y}}$ are not necessarily subsets of $\mathrm{lfp}(TP)$. \emph{The learning problem is to predict the atoms in $\mathcal{Y}$ from the atoms in $\tilde{\mathcal{X}}$ by learning from a sequence of noisy observations $(\tilde{\mathcal{X}}, \tilde{\mathcal{Y}})$, thus uncovering the underlying program $P$.}

\subsubsection{Relational Tsetlin Machine with Constants}

We base our Relational \ac{TM} on mapping the learning problem to a Propositional \ac{TM} pattern recognition problem. We consider Horn clauses without variables first. In brief, we map every atom in $HB$ to a propositional input $x_k$, obtaining the propositional input vector $X=(x_1,\ldots,x_o)$ (cf. Section~\ref{sec:TM_foundation}). That is, consider the $w$-arity relation $r_u \in R$, which takes $w$ symbols from $A$ as input. This relation can thus take $q^w$ unique input combinations. As an example, with the constants $A= \{a_1, a_2\}$ and the binary relations $R= \{r_1, r_2\}$, we get $8$ propositional inputs: $x_1^{1,1} \equiv r_1(a_1,a_1)$; $x_1^{1,2} \equiv r_1(a_1,a_2)$; $x_1^{2,1} \equiv r_1(a_2,a_1)$;$x_1^{2,2} \equiv r_1(a_2,a_2)$; $x_2^{1,1} \equiv r_2(a_1,a_1)$; $x_2^{1,2} \equiv r_2(a_1,a_2)$; $x_2^{2,1} \equiv r_2(a_2,a_1)$; and $x_2^{2,2} \equiv r_2(a_2,a_2)$. Correspondingly, we perform the same mapping to get the propositional output vector $Y$.

Finally, obtaining an input $(\tilde{\mathcal{X}}, \tilde{\mathcal{Y}})$, we set the propositional input $x_k$ to true iff its corresponding atom is in $\tilde{\mathcal{X}}$, otherwise it is set to false. Similarly, we set the propositional output variable $y_m$ to true iff its corresponding atom is in $\tilde{\mathcal{Y}}$, otherwise it is set to false.

Clearly, after this mapping, we get a Propositional \ac{TM} pattern recognition problem that can be solved as described in Section \ref{sec:TM_foundation} for a single propositional output $y_m$. This is illustrated as Step 1 in Fig. \ref{figure:tm_steps}. 

\subsubsection{Detaching the Relational \ac{TM} from Constants}

The \ac{TM} can potentially deal with thousands of propositional inputs. However, we now detach our Relational \ac{TM} from the constants, introducing Horn clauses with variables. Our intent is to provide a more compact representation of the program and to allow generalization beyond the data. Additionally, the detachment enables faster learning even with less data.

Let $\mathcal{Z} = \{Z_1, Z_2, \ldots, Z_z\}$ be $z$ variables representing the constants appearing in an observation $(\tilde{\mathcal{X}}, \tilde{\mathcal{Y}})$. Here, $z$ is the largest number of unique constants involved in any particular observation $(\tilde{\mathcal{X}}, \tilde{\mathcal{Y}})$, each requiring its own variable.

Seeking Horn clauses with variables instead of constants, we now only need to consider atoms over variable configurations (instead of over constant configurations). Again, we map the atoms to propositional inputs to construct a propositional \ac{TM} learning problem. That is, each propositional input $x_k$ represents a particular atom with a specific variable configuration: $x_k \equiv r_u(Z_{\alpha_1}, Z_{\alpha_2}, \ldots, Z_{\alpha_{w_u}})$, with $w_u$ being the arity of $r_u$. Accordingly, the number of constants in $A$ no longer affects the number of propositional inputs $x_k$ needed to represent the problem. Instead, this is governed by the number of variables in $\mathcal{Z}$ (and, again, the number of relations in $R$). That is, the number of propositional inputs is bounded by $O(z^w)$, with $w$ being the largest arity of the relations in $R$.

To detach the Relational \ac{TM} from the constants, we first replace the constants in $\tilde{\mathcal{Y}}$ with variables, from left to right. Accordingly, the corresponding constants in $\tilde{\mathcal{X}}$ is also replaced with the same variables (Step 2 in Fig. \ref{figure:tm_steps}). Finally, the constants now remaining in $\tilde{\mathcal{X}}$ is arbitrarily replaced with additional variables (Step 3 in Fig. \ref{figure:tm_steps}).

\subsubsection{Relational Tsetlin Machine Convolution over Variable Assignment Permutations}
\label{subsubsec:variablepermutaion}
Since there may be multiple ways of assigning variables to constants, the above approach may produce redundant rules. One may end up with equivalent rules whose only difference is syntactic, i.e., the same rules are expressed using different variable symbols. This is illustrated in Step 3 of Fig. \ref{figure:tm_steps}, where variables can be assigned to constants in two ways. To avoid redundant rules, the Relational \ac{TM} produces all possible permutations of variable assignments. To process the different permutations, we finally perform a convolution over the permutations in Step 4, employing a \ac{TM} convolution operator \cite{granmo2019convolutional}. The target value of the convolution is the truth value of the consequent (Step 5). 

\subsubsection{Walk-through of Algorithm with Example Learning Problem}

Fig. \ref{figure:tm_steps} contains an example of the process of detaching a Relational \ac{TM} from constants. We use the parent-grandparent relationship as an example,  employing the following Horn clause.
\[
grandparent(Z_1,Z_2) \gets parent(Z_1, Z_3),parent(Z_3,Z_2).
\]
\noindent We replace the constants in each training example with variables, before learning the clauses. Thus the Relational \ac{TM} never ``sees" the constants, just the generic variables. 

\noindent Assume the first training example is

$Input:\ parent(Bob,\ Mary)=1; $

$Target\ output:\ child(Mary,\ Bob)=1.$

\noindent Then Mary is replaced with $Z_1$ and Bob with $Z_2$ in the target output:

$Input:\ parent(Bob,\ Mary) = 1; $

$Target\ output:\ child(Z_1,\ Z_2) = 1$.

\noindent We perform the same exchange for the input, getting:

$Input:\ parent(Z_2,\ Z_1)=1; $

$Target\ output:\ child(Z_1,\ Z_2)=1.$

\noindent Here, ``parent($Z_2, Z_1$)" is treated as an input feature by the Relational \ac{TM}. That is, ``parent($Z_2, Z_1$)" is seen as a single propositional variable that is either 0 or 1, and the name of the variable is simply the string ``parent($Z_2, Z_1$)". The constants may be changing from example to example, so next time it may be Mary and Ann. However, they all end up as $Z_1$ or $Z_2$ after being replaced by the variables.
\noindent After some time, the Relational \ac{TM} would then learn the following clause:

$child(Z_1,\ Z_2)\ \gets\ parent(Z_2, Z_1).$

\noindent This is because the feature ``parent($Z_2, Z_1$)" predicts ``child($Z_1, Z_2$)" perfectly. Other candidate features like ``parent($Z_1, Z_1$)" or ``parent($Z_2, Z_3$)" are poor predictors of ``child($Z_1, Z_2$)" and will be excluded by the \ac{TM}. Here, $Z_3$ is a free variable representing some other constant, different from $Z_1$ and $Z_2$.

\noindent Then the next training example comes along:

$Input:\ parent(Bob,\ Mary)=1;$

$Target\ output:\ child(Jane,\ Bob)=0.$

\noindent Again, we start with replacing the constants in the target output with variables:

$Input:\ parent(Bob,\ Mary)=1;$

$Target\ output:\ child(Z_1,\ Z_2)=0$

\noindent which is then completed for the input:

$Input:\ parent(Z_2,\ Mary)=1; $

$Target\ output:\ child(Z_1,\ Z_2)=0.$

\noindent The constant Mary was not in the target output, so we introduce a free variable $Z_3$ for representing Mary:

$Input:\ parent(Z_2,\ Z_3)=1; $

$Target\; output:\ child(Z_1,\ Z_2)=0.$

\noindent The currently learnt clause was: 

$child(Z_1,\ Z_2)\ \gets\ parent(Z_2, Z_1).$

\noindent The feature ``parent($Z_2, Z_1$)" is not present in the input in the second training example, only ``parent($Z_2, Z_3$). Assuming a closed world, we thus have ``parent($Z_2, Z_1$)"=0.
Accordingly, the learnt clause correctly outputs 0.

\noindent For some inputs, there can be many different ways variables can be assigned to constants (for the output, variables are always assigned in a fixed order, from $Z_1$ to $Z_z$). Returning to our grandparent example in Fig. \ref{figure:tm_steps}, if we have:

$Input:\ parent(Bob,Mary);\ parent(Mary, Peter);\\\null \qquad\qquad\qquad parent(Bob,Jane)$

$Target\ output:\ grandparent(Bob,Peter).$

\noindent replacing Bob with $Z_1$ and Peter with $Z_2$, we get:

$Input:\ parent(Z_1,Mary);\ parent(Mary, Z_2);\\\null \qquad\qquad\qquad parent(Z_1,Jane)$

$Target\; output:\ grandparent(Z_1,Z_2).$

\noindent Above, both Mary and Jane are candidates for being $Z_3$. One way to handle this ambiguity is to try both, and pursue those that make the clause evaluate correctly, which is exactly how the \ac{TM} convolution operator  works \cite{granmo2019convolutional}. 

\noindent Note that above, there is an implicit existential quantifier over $Z_3$. That is, $\forall Z_1,Z_2 (\exists Z_3 (parent(Z_1,Z_3) \wedge parent(Z_3,Z_2) \rightarrow grandparent(Z_1,Z_2))$.

\begin{figure*}[ht]
\centering
\includegraphics[width=0.8\linewidth]{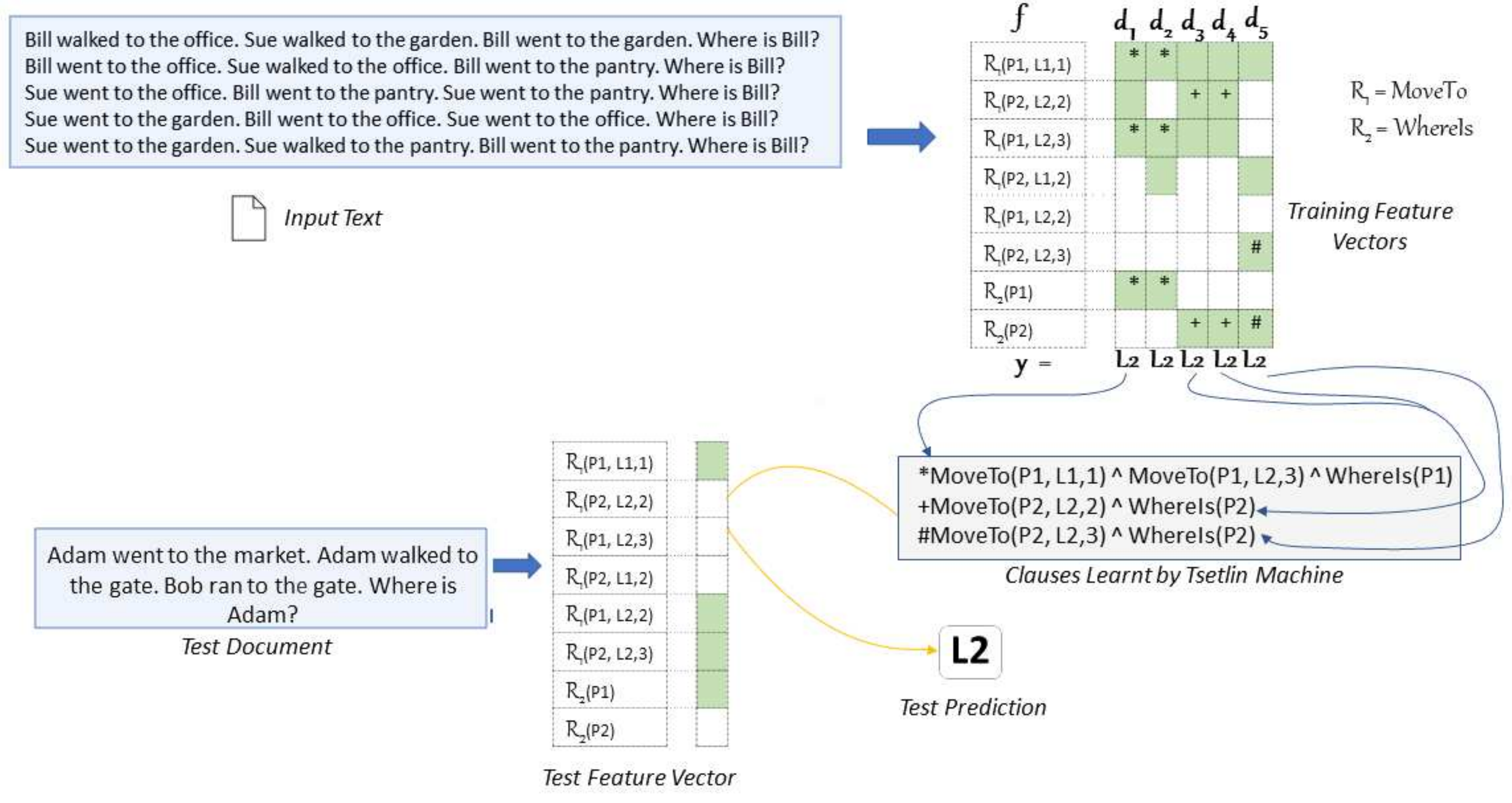}
\caption{The Relational \ac{TM} in operation}
\label{figure:tm_architecture_basic}
\end{figure*}

A practical view of how the \ac{TM} learns in such a scenario is shown in Fig. \ref{figure:tm_architecture_basic}. Continuing in the same vein as the previous examples, the Input Text in the figure is a set of statements, each followed by a question. The text is reduced to a set of relations, which act as the features for the \ac{TM} to learn from. The figure illustrates how the \ac{TM} learns relevant information (while disregarding the irrelevant), in order to successfully answer a new question (Test Document). The input text is converted into a features vector which indicates the presence or absence of relations ($R_1, R_2$) in the text, where $R_1$ and $R_2$ are respectively $MoveTo$ (in the statements) and $WhereIs$ (in the question) relations. For further simplification and compactness of representation, instead of using person and location names, those specific details are replaced by ($P_1,P_2$) and ($L_1,L_2$), respectively. In each sample, the person name that occurs first is termed $P_1$ throughout, the second unique name is termed $P_2$, and so on, and similarly for the locations. As seen in the figure, the \ac{TM} reduces the feature-set representation of the input into a set of logical conditions or clauses, all of which together describe scenarios in which the answer is $L_2$ (or Location 2). 

\textbf{Remark 1.} We now return to the implicit existential and universal quantifiers of the Horn clauses, exemplified in: $\forall Z_1,Z_2 (\exists Z_3 (parent(Z_1,Z_3) \wedge parent(Z_3,Z_2) \rightarrow grandparent(Z_1,Z_2))$. A main goal of the procedure in Fig. \ref{figure:tm_steps} is to correctly deal with the quantifiers “for all” and “exists”. ``For all" maps directly to the \ac{TM} architecture because the \ac{TM} is operating with conjunctive clauses and the goal is to make these evaluate to~$1$ (True) whenever the learning target is~$1$. ``For all" quantifiers are taken care of in Step~3 of the relational learning procedure.

\textbf{Remark 2.} ``Exists" is more difficult because it means that we are looking for a specific value for the variables in the scope of the quantifier that makes the expression evaluate to $1$. This is handled in the Steps~4-6 in Fig.~\ref{figure:tm_steps}, by evaluating all alternative values (all permutations with multiple variables involved). Some values make the expression evaluate to $0$ (False) and some make the expression become $1$. If none makes the expression~$1$, the output of the clause is~$0$. Otherwise, the output is~$1$. Remarkably, this is exactly the behavior of the \ac{TM} convolution operator defined in \cite{granmo2019convolutional}, so we have an existing learning procedure in place to deal with the ``Exists" quantifier. (If there exists a patch in the image that makes the clause evaluate to $1$, the clause evaluates to~$1$).

\section{QA in a Relational \ac{TM} Framework}
\label{sec:qa in relTM}

\begin{figure}[h]
\centering
\includegraphics[width=.45\textwidth]{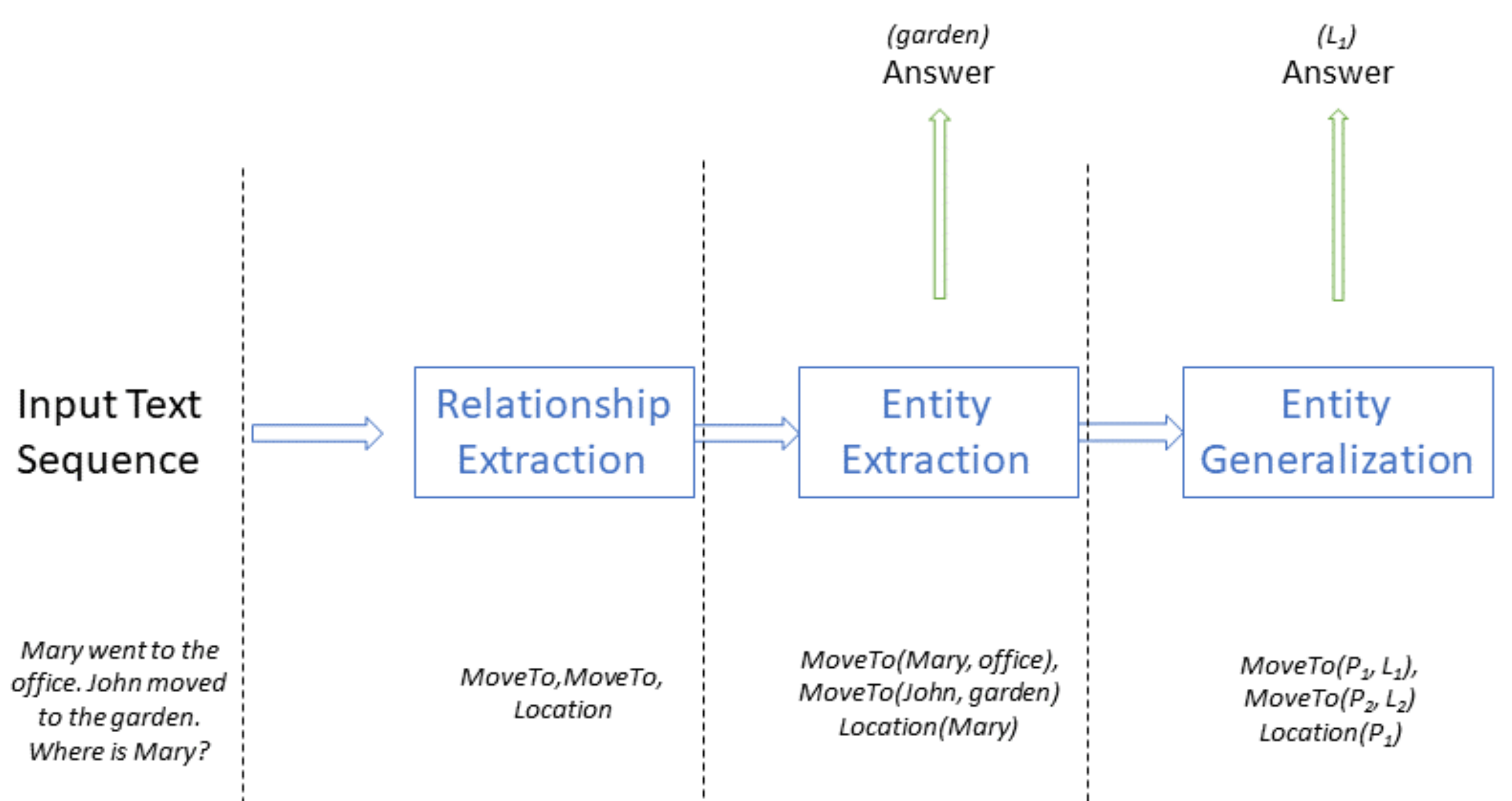}
\caption{General Pipeline} 
\label{figPipe}
\end{figure}

In this section, we describe the general pipeline to reduce natural language text into a machine-understandable relational representation to facilitate question answering. Fig. \ref{figPipe} shows the pipeline diagrammatically, with a small example. Throughout this section, we make use of two toy examples in order to illustrate the steps. One of them is derived from a standard question answering dataset \cite{weston2015towards}. The other is a simple handcrafted dataset, inspired by \cite{kowalski1979algorithm}, that refers to parent-child relationships among a group of people. Both datasets consist of instances, where each instance is a set of two or more statements, followed by a query. The expected output for each instance is the answer to the query based on the statements.

\subsection{Relation Extraction}
As a first step, we need to extract the relation(s) present in the text. A relation here is a connection between two (or more) elements of a text. As discussed before, relations occur in natural language, and reducing a text to its constituent relations makes it more structured while ignoring superfluous linguistic elements, leading to easier understanding. We assume that our text consists of simple sentences, that is, each sentence contains only one relation. The relation found in the query is either equal to, or linguistically related to the relations found in the statements preceding the query. 

Table \ref{table:RelationExtraction} shows examples of Relation Extraction on our two datasets. In Example-Movement, each statement has the relation ``MoveTo", while the query is related to ``Location". The ``Location" can be thought of as a result of the ``MoveTo" relations. Example-Parentage has ``Parent" relations as the information and ``Grandparent" as the query.

\begin{table}[!ht]
\centering
\caption{Relation Extraction}
\label{table:RelationExtraction}
\begin{tabular}{ll}
\hline
\multicolumn{1}{|l|}{\textit{Sentence}} & \multicolumn{1}{l|}{\textit{Relation}} \\ \hline
\multicolumn{1}{|l|}{Mary went to the office.} & \multicolumn{1}{l|}{{\textcolor{blue}{ \textbf{MoveTo}}}} \\ \hline
\multicolumn{1}{|l|}{John moved to the hallway} & \multicolumn{1}{l|}{{\textcolor{blue}{ \textbf{MoveTo}}}} \\ \hline
\multicolumn{1}{|l|}{Where is Mary?} & \multicolumn{1}{l|}{{\textcolor{blue}{ \textbf{Location}}}} \\ \hline
\multicolumn{2}{c}{Example-Movement}
\end{tabular}
\qquad
\begin{tabular}{ll}
\hline
\multicolumn{1}{|l|}{\textit{Sentence}}& \multicolumn{1}{l|}{\textit{Relation}} \\ \hline
\multicolumn{1}{|l|}{Bob is a parent of Mary.} & \multicolumn{1}{l|}{{\textcolor{blue}{\textbf{Parent}}}}\\ \hline
\multicolumn{1}{|l|}{Bob is a parent of Jane.} & \multicolumn{1}{l|}{{\textcolor{blue}{\textbf{Parent}}}}\\ \hline
\multicolumn{1}{|l|}{Mary is a parent of Peter}& \multicolumn{1}{l|}{{\textcolor{blue}{\textbf{Parent}}}}\\ \hline
\multicolumn{1}{|l|}{Is Bob a grandparent of Peter?} & \multicolumn{1}{l|}{{\textcolor{blue}{\textbf{Grandparent}}}} \\ \hline
\multicolumn{2}{c}{Example-Parentage}
\end{tabular}
\end{table}

\subsection{Entity Extraction}
Once the relations have been identified, the next step is to identify the elements of the text (or the entities) that are participating in the respective relations. Doing so allows us to further enrich the representation with the addition of restrictions (often real-world ones), which allow the Relational \ac{TM} to learn rules that best represent actions and their consequences in a concise, logical form. Since the datasets we are using here consist only of simple sentences, each relation is limited to having a maximum of two entities (the relations are unary or binary). 

In this step, the more external world knowledge that can be combined with the extracted entities, the richer the resultant representation. In Table \ref{table:EntityExtraction}, Example-Movement, we could add the knowledge that ``MoveTo" relation always involves a ``Person" and a ``Location". Or in Example-Parentage, ``Parent" is always between a ``Person" and a ``Person". This could, for example, prevent questions like ``Jean-Joseph Pasteur was the father of Louis Pasteur. Louis Pasteur is the father of microbiology. Who is the grandfather of microbiology?"

\noindent
Note that, as per Fig. \ref{figPipe}, it is only possible to start answering the query after both Relation Extraction and Entity Extraction have been performed, and not before. Knowledge of the Relation also allows us to narrow down possible entities for answering the query successfully. 

\begin{table}[h]
\centering
\setlength\tabcolsep{2.5pt}
\caption{Entity Extraction}
\label{table:EntityExtraction}
\begin{tabular}{lll}
\hline
\multicolumn{1}{|l|}{\textit{Sentence}} & \multicolumn{1}{l|}{\textit{Relation}} & \multicolumn{1}{l|}{\textit{Entities}} \\ \hline
\multicolumn{1}{|l|}{Mary went to the office.} & \multicolumn{1}{l|}{{\textcolor{blue}{ \textbf{MoveTo}}}} & \multicolumn{1}{l|}{{\textcolor[RGB]{0,100,0}{\textbf{Mary, office}}}} \\ \hline
\multicolumn{1}{|l|}{John moved to the hallway} & \multicolumn{1}{l|}{{\textcolor{blue}{ \textbf{MoveTo}}}} & \multicolumn{1}{l|}{{\textcolor[RGB]{0,100,0}{\textbf{John, hallway}}}} \\ \hline
\multicolumn{1}{|l|}{Where is Mary?} & \multicolumn{1}{l|}{{\textcolor{blue}{ \textbf{Location}}}} & \multicolumn{1}{l|}{{\textcolor[RGB]{0,100,0}{\textbf{Mary, ?}}}} \\ \hline
\multicolumn{3}{c}{Example-Movement}
\end{tabular}
\qquad
\setlength\tabcolsep{2.5pt}
\begin{tabular}{lll}
\hline
\multicolumn{1}{|l|}{\textit{Sentence}}& \multicolumn{1}{l|}{\textit{Relation}} & \multicolumn{1}{l|}{\textit{Entities}} \\ \hline
\multicolumn{1}{|l|}{Bob is a parent of Mary.} & \multicolumn{1}{l|}{{\textcolor{blue}{ \textbf{Parent}}}}& \multicolumn{1}{l|}{{\textcolor[RGB]{0,100,0}{\textbf{Bob, Mary}}}} \\ \hline
\multicolumn{1}{|l|}{Bob is a parent of Jane.} & \multicolumn{1}{l|}{{\textcolor{blue}{ \textbf{Parent}}}}& \multicolumn{1}{l|}{{\textcolor[RGB]{0,100,0}{\textbf{Bob, Jane}}}} \\ \hline
\multicolumn{1}{|l|}{Mary is a parent of Peter}& \multicolumn{1}{l|}{{\textcolor{blue}{ \textbf{Parent}}}}& \multicolumn{1}{l|}{{\textcolor[RGB]{0,100,0}{\textbf{Mary, Peter}}}} \\ \hline
\multicolumn{1}{|l|}{Is Bob a grandparent of Peter?} & \multicolumn{1}{l|}{{\textcolor{blue}{ \textbf{Grandparent}}}} & \multicolumn{1}{l|}{{\textcolor[RGB]{0,100,0}{\textbf{Bob, Peter}}}}\\ \hline
\multicolumn{3}{c}{Example-Parentage} \end{tabular}

\end{table}

\subsection{Entity Generalization}
\label{sec:theoryEntityGeneralization}
One of the drawbacks of the relational representation is that there is a huge increase in the number of possible relations as more and more examples are processed. One way to reduce the spread is to reduce individual entities from their specific identities to a more generalised identity. Let us consider two instances : ``Mary went to the office. John moved to the hallway. Where is Mary?" and ``Sarah moved to the garage. James went to the kitchen. Where is Sarah?". Without generalization, we end up with six different relations : MoveTo(Mary, Office), MoveTo(John, Hallway), Location(Mary), MoveTo(Sarah, Garage), MoveTo(James, Kitchen), Location(Sarah). However, to answer either of the two queries, we only need the relations pertaining to the query itself. Taking advantage of that, we can generalize both instances to just 3 relations: MoveTo(Person$_1$, Location$_1$), MoveTo(Person$_2$, Location$_2$) and Location(Person$_1$).


In order to prioritise, the entities present in the query relation are the first to be generalized. All occurrences of those entities in the relations preceding the query are also replaced by suitable placeholders.

\begin{table*}[h]
\centering
\caption{Entity Generalization : Part 1}
\begin{tabular}{lllll}
\hline
\multicolumn{1}{|l|}{\textit{Sentence}} & \multicolumn{1}{l|}{\textit{Relation}} & \multicolumn{1}{l|}{\textit{Entities}} & \multicolumn{1}{l|}{} & \multicolumn{1}{l|}{\textit{Reduced Relation}} \\ \hline
\multicolumn{1}{|l|}{Mary went to the office.} & \multicolumn{1}{l|}{{\textcolor{blue}{ \textbf{MoveTo}}}} & \multicolumn{1}{l|}{{\textcolor[RGB]{0,100,0}{\textbf{Mary, office}}}} & \multicolumn{1}{c|}{Source} & \multicolumn{1}{l|}{{\textcolor{blue}{\textbf{MoveTo}}(\textcolor{red}{X}, \color[HTML]{036400}{office})}} \\ \hline
\multicolumn{1}{|l|}{John moved to the hallway} & \multicolumn{1}{l|}{{\textcolor{blue}{ \textbf{MoveTo}}}} & \multicolumn{1}{l|}{{\textcolor[RGB]{0,100,0}{\textbf{John, hallway}}}} & \multicolumn{1}{c|}{Source} & \multicolumn{1}{l|}{{\textcolor{blue}{\textbf{MoveTo}}(\color[HTML]{036400}{John, hallway})}} \\ \hline
\multicolumn{1}{|l|}{Where is Mary?} & \multicolumn{1}{l|}{{\textcolor{blue}{ \textbf{Location}}}} & \multicolumn{1}{l|}{{\textcolor[RGB]{0,100,0}{\textbf{Mary, ?}}}} & \multicolumn{1}{l|}{Target} & \multicolumn{1}{l|}{{\textcolor{blue}{\textbf{Location}}(\textcolor{red}{X}, ?)}} \\ \hline
\multicolumn{5}{c}{Example-Movement}
\end{tabular}
\begin{tabular}{lllll}
\hline
\multicolumn{1}{|l|}{\textit{Sentence}}& \multicolumn{1}{l|}{\textit{Relation}} & \multicolumn{1}{l|}{\textit{Entities}} & \multicolumn{1}{l|}{} & \multicolumn{1}{l|}{\textit{Reduced Relation}} \\ \hline
\multicolumn{1}{|l|}{Bob is a parent of Mary.} & \multicolumn{1}{l|}{{\textcolor{blue}{ \textbf{Parent}}}}& \multicolumn{1}{l|}{{\textcolor[RGB]{0,100,0}{\textbf{Bob, Mary}}}} & \multicolumn{1}{l|}{Source} & \multicolumn{1}{l|}{{\textcolor{blue}{\textbf{Parent}}(\textcolor{red}{X}, \color[HTML]{036400}{Mary})}} \\ \hline
\multicolumn{1}{|l|}{Bob is a parent of Jane.} & \multicolumn{1}{l|}{{\textcolor{blue}{ \textbf{Parent}}}}& \multicolumn{1}{l|}{{\textcolor[RGB]{0,100,0}{\textbf{Bob, Jane}}}} & \multicolumn{1}{l|}{Source} & \multicolumn{1}{l|}{{\textcolor{blue}{\textbf{Parent}}(\textcolor{red}{X}, \color[HTML]{036400}{Jane})}} \\ \hline
\multicolumn{1}{|l|}{Mary is a parent of Peter}& \multicolumn{1}{l|}{{\textcolor{blue}{ \textbf{Parent}}}}& \multicolumn{1}{l|}{{\textcolor[RGB]{0,100,0}{\textbf{Mary, Peter}}}} & \multicolumn{1}{l|}{Source} & \multicolumn{1}{l|}{{\textcolor{blue}{\textbf{Parent}}(\color[HTML]{036400}{Mary}, \textcolor{red}{Y})}} \\ \hline
\multicolumn{1}{|l|}{Is Bob a grandparent of Peter?} & \multicolumn{1}{l|}{{\textcolor{blue}{ \textbf{Grandparent}}}} & \multicolumn{1}{l|}{{\textcolor[RGB]{0,100,0}{\textbf{Bob, Peter}}}}& \multicolumn{1}{l|}{Target} & \multicolumn{1}{l|}{{\textcolor{blue}{\textbf{Grandparent}}(\textcolor{red}{X, Y})}} \\ \hline
\multicolumn{5}{c}{Example-Parentage}
\end{tabular}
\end{table*}


The entities present in the other relations are then replaced by what can be considered as free variables (since they do not play a role in answering the query).

In the next section we explain how this relational framework is utilized for question answering using \ac{TM}.
\begin{table*}[h]
\centering
\caption{Entity Generalization : Part 2}
\begin{tabular}{lllll}
\hline
\multicolumn{1}{|l|}{\textit{Sentence}} & \multicolumn{1}{l|}{\textit{Relation}} & \multicolumn{1}{l|}{\textit{Entities}} & \multicolumn{1}{l|}{} & \multicolumn{1}{l|}{\textit{Reduced Relation}} \\ \hline
\multicolumn{1}{|l|}{Mary went to the office.} & \multicolumn{1}{l|}{{\textcolor{blue}{ \textbf{MoveTo}}}} & \multicolumn{1}{l|}{{\textcolor[RGB]{0,100,0}{\textbf{Mary, office}}}} & \multicolumn{1}{c|}{Source} & \multicolumn{1}{l|}{{\textcolor{blue}{\textbf{MoveTo}}(\textcolor{red}{X, A})}} \\ \hline
\multicolumn{1}{|l|}{John moved to the hallway} & \multicolumn{1}{l|}{{\textcolor{blue}{ \textbf{MoveTo}}}} & \multicolumn{1}{l|}{{\textcolor[RGB]{0,100,0}{\textbf{John, hallway}}}} & \multicolumn{1}{c|}{Source} & \multicolumn{1}{l|}{{\textcolor{blue}{\textbf{MoveTo}}(Y, B}) } \\ \hline
\multicolumn{1}{|l|}{Where is Mary?} & \multicolumn{1}{l|}{{\textcolor{blue}{ \textbf{Location}}}} & \multicolumn{1}{l|}{{\textcolor[RGB]{0,100,0}{\textbf{Mary, ?}}}} & \multicolumn{1}{l|}{Target} & \multicolumn{1}{l|}{{\textcolor{blue}{\textbf{Location}}(\textcolor{red}{X}, ?)}} \\ \hline
\multicolumn{5}{c}{Example-Movement}
\end{tabular}
\begin{tabular}{lllll}
\hline
\multicolumn{1}{|l|}{\textit{Sentence}}& \multicolumn{1}{l|}{\textit{Relation}} & \multicolumn{1}{l|}{\textit{Entities}} & \multicolumn{1}{l|}{} & \multicolumn{1}{l|}{\textit{Reduced Relation}} \\ \hline
\multicolumn{1}{|l|}{Bob is a parent of Mary.} & \multicolumn{1}{l|}{{\textcolor{blue}{ \textbf{Parent}}}}& \multicolumn{1}{l|}{{\textcolor[RGB]{0,100,0}{\textbf{Bob, Mary}}}} & \multicolumn{1}{l|}{Source} & \multicolumn{1}{l|}{{\textcolor{blue}{\textbf{Parent}}(\textcolor{red}{X, Z})} } \\ \hline
\multicolumn{1}{|l|}{Bob is a parent of Jane.} & \multicolumn{1}{l|}{{\textcolor{blue}{ \textbf{Parent}}}}& \multicolumn{1}{l|}{{\textcolor[RGB]{0,100,0}{\textbf{Bob, Jane}}}} & \multicolumn{1}{l|}{Source} & \multicolumn{1}{l|}{{\textcolor{blue}{\textbf{Parent}}(\textcolor{red}{X}, W)}}\\ \hline
\multicolumn{1}{|l|}{Mary is a parent of Peter}& \multicolumn{1}{l|}{{\textcolor{blue}{ \textbf{Parent}}}}& \multicolumn{1}{l|}{{\textcolor[RGB]{0,100,0}{\textbf{Mary, Peter}}}} & \multicolumn{1}{l|}{Source} & \multicolumn{1}{l|}{{\textcolor{blue}{\textbf{Parent}}(\textcolor{red}{Z, Y})}} \\ \hline
\multicolumn{1}{|l|}{Is Bob a grandparent of Peter?} & \multicolumn{1}{l|}{{\textcolor{blue}{ \textbf{Grandparent}}}} & \multicolumn{1}{l|}{{\textcolor[RGB]{0,100,0}{\textbf{Bob, Peter}}}}& \multicolumn{1}{l|}{Target} & \multicolumn{1}{l|}{{\textcolor{blue}{\textbf{Grandparent}}(\textcolor{red}{X, Y})}} \\ \hline
\multicolumn{5}{c}{Example-Parentage}
\end{tabular}
\end{table*}

\subsection{Computational Complexity in Relation \ac{TM} }
\label{subsec:computationalcomplexity}
One of the primary differences between the relational framework proposed in this paper versus the existing \ac{TM} framework lies in Relation Extraction and Entity Generalization. 
The reason for these steps is that they allow us more flexibility in terms of what the \ac{TM} learns, while keeping the general learning mechanism unchanged. 

Extracting relations allows the \ac{TM} to focus only on the major operations expressed via language, without getting caught up in multiple superfluous expressions of the same thing. It also enables to bridge the gap between structured and unstructured data. Using relations helps the resultant clauses be closer to real-world phenomena, since they model actions and consequences, rather than the interactions between words. 

Entity Generalization allows the clauses to be succinct and precise, adding another layer of abstraction away from specific literals, much like Relation Extraction. It also gives the added benefit of making the learning of the \ac{TM} more generalized. In fact, due to this process, the learning reflects `concepts' gleaned from the training examples, rather than the examples themselves. 

To evaluate computational complexity, we employ the three
costs $\alpha$, $\beta$, and $\gamma$ , where in terms of computational cost, $\alpha$ is cost to perform the conjunction of two bits, $\beta$ is cost of computing the summation of two integers, and $\gamma$ is cost to update the state of a single
automaton. In a Propositional \ac{TM}, the worst case scenario would be the one in which all the clauses are updated. Therefore, a \ac{TM} with $m$ clauses and an input vector of o features, needs to perform $(2o+1) \times m$ number of \ac{TA} updates for a single training sample. For a total of $d$ training samples, the total cost of updating is $d \times \gamma \times (2o+1) \times m$. 

Once weight updates have been successfully performed, the next step is to calculate the clause outputs. Here the worst case is represented by all the clauses including all the corresponding literals, giving us a cost of $\alpha \times 2o \times m$ (for a single sample). 

The last step involves calculating the difference in votes from the clause outputs, thus incurring a per-sample cost of $\beta \times (m - 1)$.

Taken together, the total cost function for a Propositional \ac{TM} can be expressed as:

$f(d) = d \times [(\gamma \times (2o+1) \times m) + (\alpha \times 2o \times m) + (\beta \times (m - 1))] $

Expanding this calculation to the Relation \ac{TM} scenario, we need to account for the extra operations being performed, as detailed earlier: Relation Extraction and Entity Generalization.
The number of features per sample is restricted by the number of possible relations, both in the entirety of the training data, as well as only in the context of a single sample. For example, in the experiments involving ``MoveTo" relations, we have restricted our data to have 3 statements, followed by a question (elaborated further in the next section). Each statement gives rise to a single ``MoveTo" relation, which has 2 entities (a location and a person). 

When using the textual constants (i.e., without Entity Generalization), the maximum number of possible features thus becomes equal to the number of possible combination between the unique entities in the relations. Thus if each sample contains $r$ Relations, and a Relation $R$ involves $e$ different entities ($E_1, E_2, ..., E_e$), and cardinality of sets $E_1, E_2, ..., E_e$ be represented as $\lvert E_1 \rvert, \lvert E_2 \rvert, ..., \lvert E_e \rvert$, the number of features in the above equation can be re-written as 

$o = \{ \binom{\lvert E_1 \rvert}{1} \times \binom{\lvert E_2 \rvert}{1} \times ... \binom{\lvert E_e \rvert}{1}\} \times r$.

As discussed earlier in Section \ref{sec:theoryEntityGeneralization} as well as shown in the previous paragraph, this results in a large $o$, since it depends on the number of unique textual elements in each of the entity sets. Using Entity Generalization, the number of features no longer depends on the cardinality of set $E_{n, 1\leq n\leq e}$ in the context of the whole dataset. Instead, it only depends on the context of the single sample. Thus, if each sample contains $r$ Relations, and a Relation $R$ involves $e$ different entities ($E_1, E_2, ..., E_e$), and maximum possible cardinality of sets $E_1, E_2, ..., E_e$ are $\lvert E_1 \rvert = \lvert E_2 \rvert = ... = \lvert E_n \rvert = r$, the number of features become

$o = \{ \binom{\lvert r \rvert}{1} \times \binom{\lvert r \rvert}{1} \times ... \binom{\lvert r \rvert}{1}\} \times r = r^{(e+1)}$. 

In most scenarios, this number is much lower than the one obtained without the use of Entity Generalization.

A little further modification is required when using the convolutional approach. In calculating $f(d)$, the measure of $o$ remains the same as we just showed with/without Entity Generalization. However the second term in the equation, which refers to the calculation of clause outputs ($d \times \alpha \times 2o \times m$), changes due to the difference in mechanism of calculating outputs for convolutional and non-convolutional approaches. In the convolutional approach, with Entity Generalization, we need to consider free and bound variables in the feature representation. Bound variables are the ones which are linked by appearance in the source and target relations, while the other variables, which are independent of that restriction, can be referred to as the free variables. Each possible permutation of the free variables in different positions are used by the convolutional approach to determine the best generic rule that describes the scenario. In certain scenarios, it may be possible to have certain permutations among the bound variables as well, without violating the restrictions added by the relations. One such scenario is detailed with an example in Section \ref{sec:experimentEntityPermuation}, where a bound ``Person'' entity can be permuted to allow any other ``Person'' entity, as long as the order of ``MoveTo'' relations are not violated. However, it is difficult to get a generic measure for the same, which would work irrespective of the nature of the relation (or their restrictions). Therefore, for the purpose of this calculation, we only take into account the permutations afforded to us by the free variable. Using the same notation as before, if each sample contains $r$ Relations, and a Relation $R$ involves $e$ different entities, the total number of variables is $R \times e$. Of these, if $v$ is the number of free variables, then they can be arranged in $v!$ different ways. Assuming $v$ is constant for all samples, the worst case $f(d)$ can thus be rewritten as 

$f(d) = d \times [(\gamma \times (2o+1) \times m) + (v! \times \alpha \times 2o \times m) + (\beta \times (m - 1))] $ .

\section{Experimental study}
\label{sec:experiments}
To further illustrate how the \ac{TM} based logic modelling works practically, we employ examples from a standard question answering dataset \cite{weston2015towards}. For the scope of this work, we limit ourselves to the first subtask as defined in the dataset, viz. a question that can be answered by the preceding context, and the context contains a single supporting fact. 

To start with, there is a set of statements, followed by a question, as discussed previously. For this particular subtask, the answer to the question is obtained by a single statement out of the set of statements provided (hence the term, single supporting fact). 

\textbf{Input:}\textit{William moved to the office. Susan went to the garden. William walked to the pantry. Where is William?
}

\textbf{Output:} pantry

\noindent We assume the following knowledge to help us construct the task:
\begin{enumerate}
\item All statements only contain information pertaining to relation MoveTo
\item All questions only relate to information pertaining to relation CurrentlyAt
\item Relation MoveTo involves 2 entities, such that {$MoveTo(a,b) : a \in \{Persons\}, b\in \{Locations\}$}
\item Relation CurrentlyAt involves 2 entities, such that {$CurrentlyAt(a,b) : a \in \{Persons\}, b\in \{Locations\}$}
\item {MoveTo is a time-bound relation, it's effect is superseded by a similar action performed at a later time.}
\end{enumerate}

\subsection{Without Entity Generalization}

The size of the set of statements from which the model has to identify the correct answer influences the complexity of the task. For the purpose of this experiment, the data is capped to have a maximum of three statements per input, and over all has five possible locations. This means that the task for the \ac{TM} model is reduced to classifying the input into one of five possible classes.

To prepare the data for the \ac{TM}, the first step involves reducing the input to relation-entity bindings. These bindings form the basis of our feature set, which is used to train the \ac{TM}. Consider the following input example:

\textbf{Input} $=>$ MoveTo(William, Office), MoveTo(Susan, Garden), MoveTo(William, Pantry), Q(William).

Since the \ac{TM} requires binary features, each input is converted to a vector, where each element represents the presence (or absence) of the relationship instances.

Secondly, the list of possible answers is obtained from the data, which is the list of class labels. Continuing our example, possible answers to the question could be:

\textbf{Possible Answers} : [Office, Pantry, Garden, Foyer, Kitchen]

Once training is complete, we can use the inherent interpretability  of the \ac{TM} to get an idea about how the model learns to discriminate the information given to it. The set of all clauses arrived at by the \ac{TM} at the end of training represents a global view of the learning, i.e. what the model has learnt in general. The global view can also be thought of as a description of the task itself, as understood by the machine. We also have access to a local snapshot, which is particular to each input instance. The local snapshot involves only those clauses that help in arriving at the answer for that particular instance. 

Table \ref{table_local_snap_eg} shows the local snapshot obtained for the above example. As mentioned earlier, the \ac{TM} model depends on two sets of clauses for each class, a positive set and a negative set. The positive set represents information favour of the class, while the negative set represents the opposite. The sum of the votes given by these two sets thus represent the final class the model decides on. As seen in the example, all the classes other than ``Pantry" receive more negative votes than positive, making it the winning class. The clauses themselves allow us to peek into the learning mechanism. For the class ``Office", a clause captures the information that (a) the question contains ``William", and (b) the relationship MoveTo(William, Office) is available. This clause votes in support of the class, i.e. this is an evidence that the answer to the question ``Where is William?" may be ``Office". However, another clause encapsulates the previous two pieces of information, as well as something more : (c) the relationship MoveTo(William, Pantry) is available. Not only does this clause vote against the class ``Office", it also ends up with a larger share of votes than the clause voting positively. 

\begin{table*}
\centering
\caption{Local Snapshot of Clauses for example ``William moved to the office. Susan went to the garden. William walked to the pantry. Where is William?"}
\label{table_local_snap_eg}
\begin{tabular}{|c|l|c|c|c|} 
\hline
 \textbf{Class}& \textbf{Clause}& \textbf{+/-}& \multicolumn{1}{l|}{\textbf{Votes} } & \begin{tabular}[c]{@{}c@{}}\textbf{Total Votes}\\\textbf{for Class} \end{tabular} \\ 
\hline
\multirow{2}{*}{Office} & Q(William) AND MoveTo(William, Office) & + & 12 & \multirow{2}{*}{-35} \\ 
\cline{2-4}
 & Q(William) AND MoveTo(William, Office)\textasciitilde{} AND MoveTo(William, Pantry) & - & 47 &\\ 
\hline
\multirow{2}{*}{Pantry} & Q(William) AND MoveTo(William, Office)\textasciitilde{} AND MoveTo(William, Pantry) & + & 64 & \multirow{2}{*}{\textbf{+49} } \\ 
\cline{2-4}
 & Q(William) AND MoveTo(William, Office) & - & 15 &\\ 
\hline
\multirow{2}{*}{Garden} & MoveTo(Susan, Garden) & + & 12 & \multirow{2}{*}{-36} \\ 
\cline{2-4}
 & Q(William) AND MoveTo(William, Office) AND MoveTo(William, Pantry) & - & 48 &\\ 
\hline
\multirow{2}{*}{Foyer} & - & + & 0 & \multirow{2}{*}{-106} \\ 
\cline{2-4}
 & \begin{tabular}[c]{@{}l@{}}Q(William) AND MoveTo(William, Office) AND MoveTo(William, Pantry)\\AND MoveTo(Susan, Garden) \end{tabular} & - & 106 &\\ 
\hline
\multirow{2}{*}{Kitchen} & - & + & 0 & \multirow{2}{*}{-113} \\ 
\cline{2-4}
 & \begin{tabular}[c]{@{}l@{}}Q(William) AND MoveTo(William, Office) AND MoveTo(William, Pantry)\\AND MoveTo(Susan, Garden) \end{tabular} & - & 113 &\\
\hline
\end{tabular}
\end{table*}

The accuracy obtained over 100 epochs for this experiment was $94.83\%$, with a F-score of $94.80$.

\subsubsection{Allowing Negative Literals in Clauses}Above results were obtained by only allowing positive literals in the clauses. The descriptive power of the \ac{TM} goes up if negative literals are also added. However, the drawback to that is, while the \ac{TM} is empowered to make more precise rules (and by extension, decisions), the overall complexity of the clauses increase, making them less readable. Also, previously, the order of the MoveTo action could be implied by the order in which they appear in the clauses, since only one positive literal can be present per sentence, but in case of negative literals, we need to include information about the sentence order. Using the above example again, if we do allow negative literals the positive evidence for Class Office looks like:
\begin{sloppypar}
\begin{small}
$Q(William)\ AN\!D\ Not(Q(Susan))\ AN\!D\\ MoveTo(S1,William,Of\!fice)\ AN\!D\\ NOT(MoveTo(S3,William,Garden))\ AN\!D\\ NOT(MoveTo(S3,William,Foyer))\ AN\!D\\ NOT(MoveTo(S3,William,Kitchen))\ 
AN\!D\\ NOT(MoveTo(S3,Susan,Garden))\ AN\!D\\ NOT(MoveTo(S3,Susan,Of\!fice))\ AN\!D\\ NOT(MoveTo(S3,Susan,Pantry))\ AN\!D\\ NOT(MoveTo(S3,Susan,Foyer))\ AN\!D\\ NOT(MoveTo(S3,Susan,Kitchen))$.
\end{small}
\end{sloppypar}

At this point, we can see that the use of constants lead to a large number of repetitive information in terms of the rules learnt by the \ac{TM}. Generalizing the constants as per their entity type
can prevent this.

\subsection{Entity Generalization}
\label{sec:experimentEntityPermuation}
Given a set of sentences and a following question, the first step remains the same as in the previous subsection, i.e. reducing the input to relation-entity bindings. In the second step, we carry out a grouping by entity type, in order to generalize the information. Once the constants have been replaced by general placeholders, we continue as previously, collecting a list of possible outputs (to be used as class labels), and further, training a \ac{TM} based model with binary feature vectors.

As before, the data is capped to have a maximum of three statements per input. Continuing with the same example as above,

\textbf{Input:}\textit{William moved to the office. Susan went to the garden. William walked to the pantry. Where is William?
}

\textbf{Output:} pantry

\textbf{1. Reducing to relation-entity bindings:} Input $=>$ MoveTo(William, Office), MoveTo(Susan, Garden), MoveTo(William, Pantry), Q(William)

\textbf{2. Generalizing bindings:} $=>$ MoveTo(Per1, Loc1), MoveTo(Per2, Loc2), MoveTo(Per1, Loc3), Q(Per1)

\textbf{3. Possible Answers} : [Loc1, Loc2, Loc3]

\noindent The simplifying effect of generalization is seen right away: even though there are 5 possible location in the whole dataset, for any particular instance there are always maximum of three possibilities, since there are maximum three statements per instance.

\begin{table*}
\centering
\caption{Clause snapshot for ``William moved to the office. Susan went to the garden. William walked to the pantry. Where is William?" after  generalization}
\label{table_local_snap_gen_eg}
\begin{tabular}{|c|l|c|c|c|} 
\hline
\textbf{Class} & \textbf{Clause} & \textbf{+/-} & \multicolumn{1}{l|}{\textbf{Votes}} & \begin{tabular}[c]{@{}l@{}}\textbf{Total Votes}\\\textbf{for Class}\end{tabular} \\ 
\hline
\multirow{2}{*}{Loc1} & Q(Per1) AND MoveTo(Per1, Loc1) & + & 3 & \multirow{2}{*}{-44} \\ 
\cline{2-4}
 & Q(Per1) AND MoveTo(Per1, Loc1)AND MoveTo(Per1, Loc3) & - & 47 &\\ 
\hline
\multirow{2}{*}{Loc2} &MoveTo(Per1, Loc1) & + & 2 & \multirow{2}{*}{-88} \\ 
\cline{2-4}
 & Q(Per1) AND MoveTo(Per1, Loc3) & - & 90 &\\ 
\hline
\multirow{2}{*}{Loc3} & Q(Per1) AND MoveTo(Per1, Loc1)AND MoveTo(Per1, Loc3) & + & 51 & \multirow{2}{*}{\textbf{+51}} \\ 
\cline{2-4}
 & - & - & 0 &\\ 
\hline
\end{tabular}
\end{table*}

As seen from the local snapshot (Table \ref{table_local_snap_gen_eg}), the clauses formed are much more compact and easily understandable. The generalization also releases the \ac{TM} model from the restriction of having had to seen definite constants before in order to make a decision. The model can process ``Rory moved to the conservatory. Rory went to the cinema. Cecil walked to the school. Where is Rory?", without needing to have encountered constants ``Rory", ``Cecil", ``school", ``cinema" and ``conservatory". 

Accuracy for this experiment over 100 epochs was $99.48\%$ , with a F-score of $92.53$.

A logic based understanding of the relation ``Move" could typically be expressed as :

\begin{sloppypar}
\begin{small}
$\rightarrow MoveTo(William, of\!fice) + MoveTo(Susan, garden) + MoveTo(William, pantry) $ 

$\rightarrow MoveTo(P_1, of\!fice) +MoveTo(P_2, garden) +\\ MoveTo(P_1, pantry)$

$\rightarrow MoveTo(P_1, L_1) + MoveTo(P_2, L_2) + MoveTo(P_1, L_3)$

$\rightarrow MoveTo(P_1, L_1) + * + MoveTo(P_1, L_n)$

$\implies Location(P_1, L_n)$.
\end{small}
\end{sloppypar}
From the above two subsections, we can see that with more and more generalization, the learning encapsulated in the \ac{TM} model can approach what could possibly be a human-level understanding of the world.

\subsection{Variable Permutation and Convolution}
As described in Section \ref{subsubsec:variablepermutaion}, we can produce all possible permutations of the available variables in each sample (after Entity Generalization) as long as the relation constraints are not violated. Doing this gives us more information per sample:

\textbf{Input:}\textit{William moved to the office. Susan went to the garden. William walked to the pantry. Where is William?
}

\textbf{Output:} pantry

\textbf{1. Reducing to relation-entity bindings:} Input $=>$ MoveTo(William, Office), MoveTo(Susan, Garden), MoveTo(William, Pantry), Q(William)

\textbf{2. Generalizing bindings:} $=>$ MoveTo(Per1, Loc1), MoveTo(Per2, Loc2), MoveTo(Per1, Loc3), Q(Per1)

\textbf{3. Permuted Variables:} $=>$ MoveTo(Per2, Loc1), MoveTo(Per1, Loc2), MoveTo(Per2, Loc3), Q(Per2)

\textbf{4. Possible Answers} : [Loc1, Loc2, Loc3]

\noindent This has two primary benefits. Firstly, in a scenario where the given data does not encompass all possible structural differences in which a particular information maybe represented, using the permutations allows the \ac{TM} to view a closer-to-complete representation from which to build it's learning (and hence, explanations). Moreover, since the \ac{TM} can learn different structural permutations from a single sample, it ultimately requires fewer clauses to learn effectively. In our experiments, permutations using Relational \ac{TM} Convolution allowed for up to 1.5 times less clauses than using a non-convolutional approach.

As detailed in Section \ref{subsec:computationalcomplexity}, convolutional and non-convolutional approaches have different computational complexity. Hence, the convolutional approach makes sense only when the reduction in complexity from fewer clauses balance the increase due to processing the convolutional window itself.

\subsection{Noise Tolerance}
To verify our claims of noise tolerance as shown by the \ac{TM} based architecture, the above experiments were repeated, but with increasing amount of noise artificially introduced into the training data. The results are shown in Table \ref{tab:accvserror}. We observe that with 1\%, 2\%, 5\% and 10\% of noise, the testing accuracy fell by approximately 1.1\% each time when entity generalization was used. 

\begin{table}
\centering
\caption{Average Accuracy on Test with Increase in Error in Training Data}
\label{tab:accvserror}
\begin{tabular}{|l|l|l|l|l|l|} 
\hline
Error Rate & 0\% & 1\% & 2\% & 5\% & 10\% \\ 
\hline
Accuracy & 99.48 & 98.79 & 98.24 & 97.02 & 95.08 \\
\hline
\end{tabular}
\end{table}

\subsection{Horn Clause Representation}

The example elaborated in the previous section can be formulated as the following Horn clause representation:

\begin{small}
\begin{sloppypar}
\begin{enumerate}
\item $Person(Susan).$
\item $Person(William).$
\item $Location(Of\!fice).$
\item $Location(Garden).$
\item $Location(Pantry).$
\item $CurrentlyAt(Susan,Pantry).$
\item $CurrentlyAt(William,Pantry).$
\item $MoveTo(Susan, Garden) \gets Person(Susan),\\\null\; ~~~~~Location(Garden),
\\\null\;~~~~~not\ CurrentlyAt(Susan, Garden)$
\item $MoveTo(William, Of\!fice) \gets Person(William),\\\null\; ~~~~~Location(Of\!fice), 
\\\null\; ~~~~~not\ CurrentlyAt(William, Of\!fice). $
\end{enumerate}
\end{sloppypar}
\end{small}

\noindent
After generalization, we substitute ground rules 8 and 9 with the following rule:

\begin{small}
\begin{sloppypar}
\begin{enumerate}
\item[10)] $MoveTo(P, L) \gets Person(P), Location(L),
\\\null ~~~~~~not~CurrentlyAt(P, L).$
\end{enumerate}
\end{sloppypar}
\end{small}

The above set of Horn clauses define the immediate consequences operator whose LFP represents the Herbrand interpretation of our QA framework.

\section{Conclusion}
\label{conclusion}
Making interpretable logical decisions in question answering system is an area of active research. In this work, we propose a novel relational logic based \ac{TM} framework to approach QA tasks systematically. Our proposed method takes advantage of noise tolerance showed by \acp{TM} to work in uncertain or ambiguous contexts. We reduce the context-question-answer tuples to a set of logical arguments, which is used by the \ac{TM} to determine rules that mimic real-world actions and consequences. 

The resulting TM is relational (as opposed to the previously propositional TM) and can take work on logical structures that occur in natural language in order to encode rules representing actions and effects in the form of Horn clauses. We show initial results using the Relational TM on artificial datasets of closed-domain question answering, and those results are extremely promising. The use of first-order representations, as described in this paper, allows KBs to be up to $10$ times smaller, while at the same time showing an answering accuracy increase of almost $5\%$ to $99.48\%$.

Further work on this framework will involve a larger number of relations, with greater inter-dependencies, and analyzing how well the \ac{TM} can learn the inherent logical structure governing such dependencies. We also intend to introduce recursive Horn clauses to make the computing power of the Relational \ac{TM}  equivalent to a universal Turing machine. Moreover, we wish to experiment with this framework on real-world natural language datasets, rather than on toy ones. A prominent example is exploring large corpus of documents related to human rights violation and using them to assess risks of social instability. We expect that the resultant logic structures will be large and complicated, however, once obtained, can be used to effectively translate to and fro between the machine world and the real world.



%
\bibliographystyle{IEEEtran}
\bibliography{references}

%
\begin{IEEEbiography}[{\includegraphics[width=1in,height=1.25in,clip,keepaspectratio]{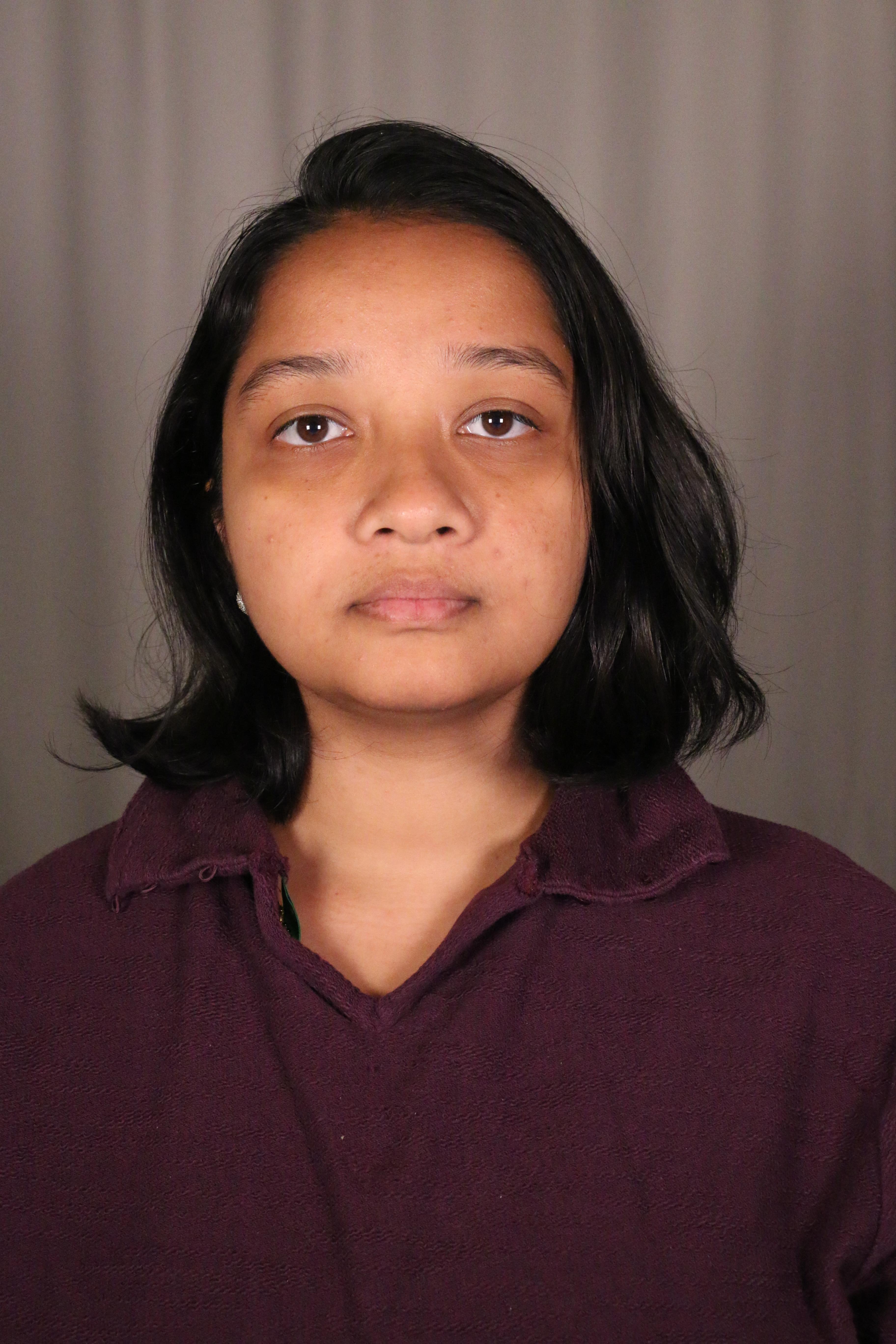}}]{Rupsa Saha}

received her M.Tech degree in information and communication technology with specialization in machine
intelligence from DAIICT, India in 2014. She is currently pursuing her Ph.D. on Tsetlin Machines with the Centre for Artificial Intelligence Research, University of Agder, Norway. Her research interests include machine learning, NLP and chatbots.
\end{IEEEbiography}

\begin{IEEEbiography}[{\includegraphics[width=1in,height=1.25in,clip,keepaspectratio]{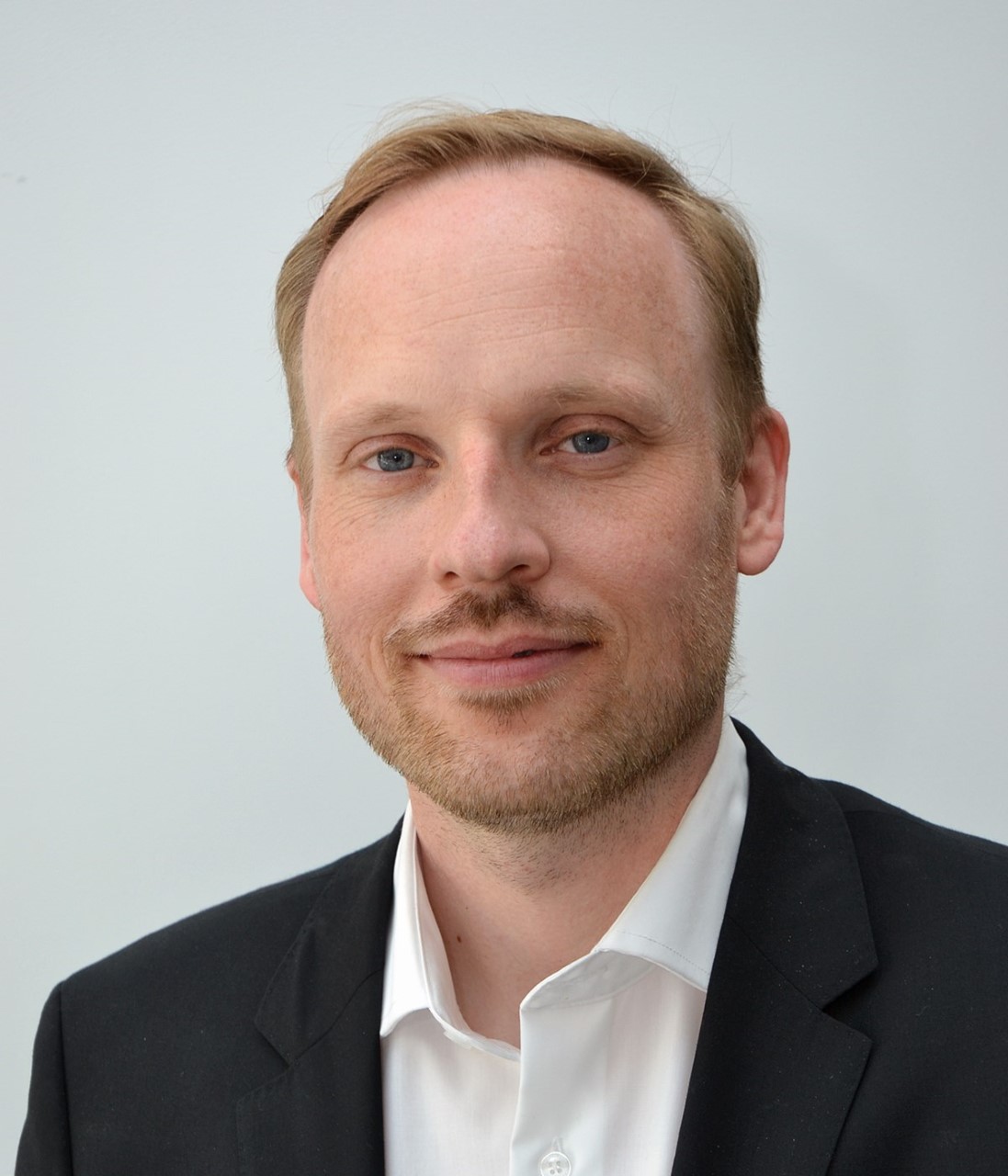}}]{Ole-Christoffer Granmo}
is a Professor and Founding Director of Centre for Artificial Intelligence Research (CAIR), University of Agder, Norway. He obtained his master’s degree in 1999 and the PhD degree in 2004, both from the University of Oslo, Norway. Dr. Granmo has authored in excess of 140 refereed papers with 6 best paper awards, encompassing learning automata, bandit algorithms, Tsetlin machines, Bayesian reasoning, reinforcement learning, and computational linguistics. He has further coordinated 7+ Norwegian Research Council projects and graduated more than 60 master- and PhD students. Dr. Granmo is also a co-founder of the Norwegian Artificial Intelligence Consortium (NORA). Apart from his academic endeavours, he co-founded the company Anzyz Technologies AS.
\end{IEEEbiography}

\begin{IEEEbiography}[{\includegraphics[width=1in,height=1.25in,clip,keepaspectratio]{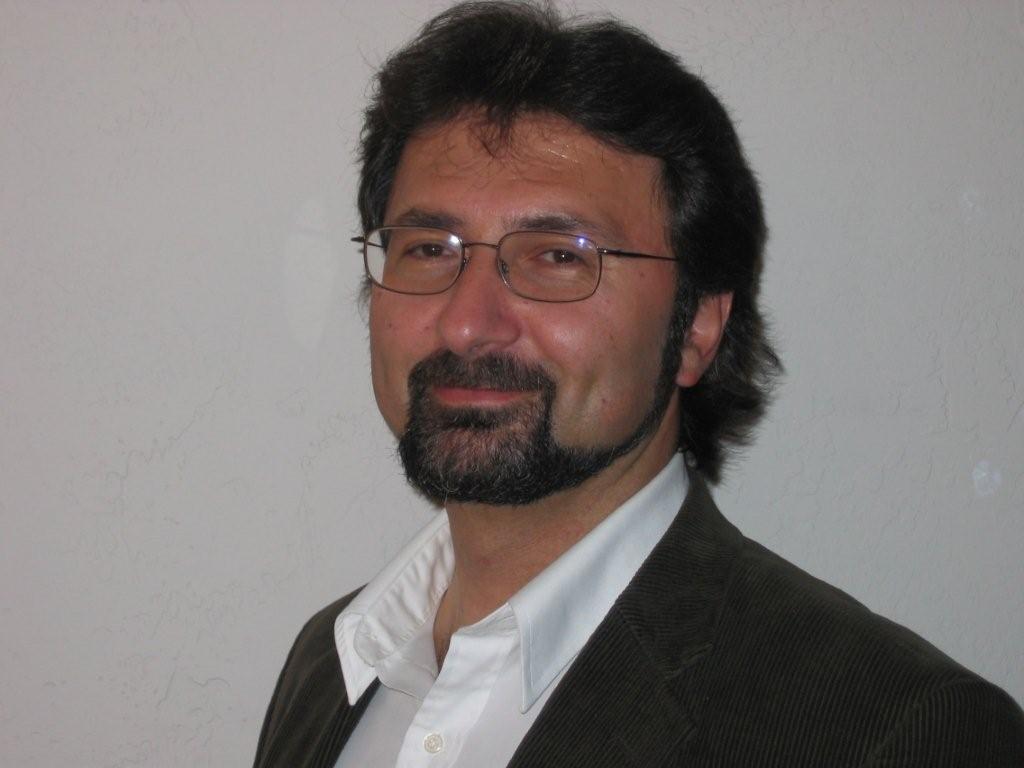}}]{Vladimir I. Zadorozhny} is a Professor at the University of Pittsburgh School of Computing and Information.  He is also a Core Faculty Member at the University of Pittsburgh Biomedical Informatics Training Program, an Adjunct Professor at Faculty of Engineering and Science and a member of the Centre for Artificial Intelligence Research, University of Agder, Norway. He received his Ph.D. in 1993 from the Institute for Problems of Informatics, Russian Academy of Sciences in Moscow. Before coming to USA in 1998 he was a Principal Research Scientist in the Institute of System Programming, Russian Academy of Sciences. His research interests include information integration, data fusion, complex adaptive systems and scalable architectures for wide-area environments. He specifically interested in application of scalable data fusion methods to enable efficient data processing and sense-making in complex domains. His research has been supported by NSF, EU and Norwegian Research Council. Vladimir is a recipient of Fulbright Scholarship for 2014-2015.  He has received several best paper awards and has chaired and served on program committees of multiple Database and Distributed Computing Conferences and Workshops.
\end{IEEEbiography}

\begin{IEEEbiography}[{\includegraphics[width=1in,height=1.25in,clip,keepaspectratio]{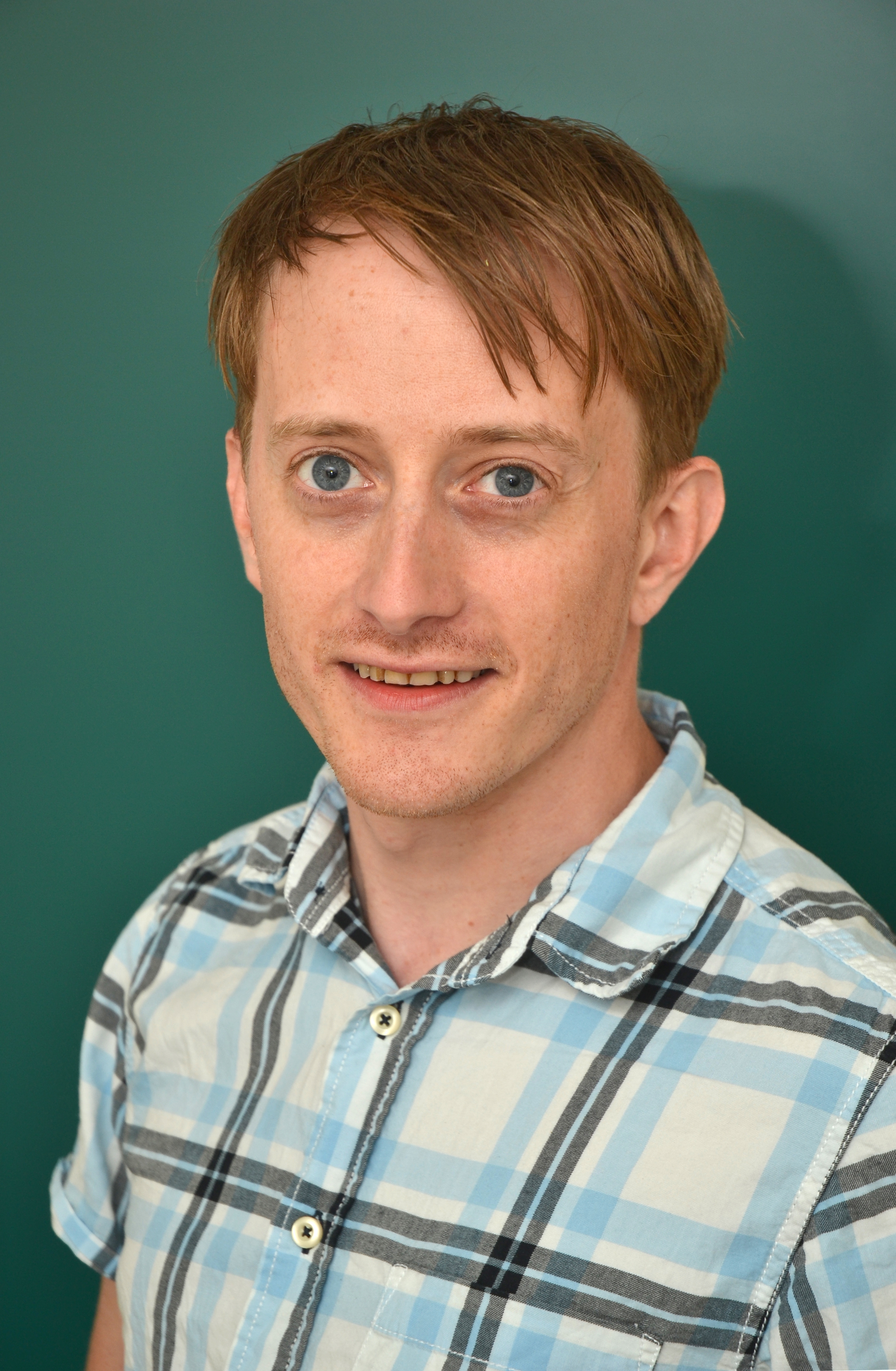}}]{Morten Goodwin}
received the B.Sc. and M.Sc. degrees from the University of Agder, Norway, in 2003 and 2005, respectively, and the Ph.D. degree from Aalborg University Department of Computer Science, Denmark, in 2011, on applying machine learning algorithms on eGovernment indicators which are difficult to measure automatically.
He is a Professor with the Department of ICT, the University of Agder, deputy director for Centre for Artificial Intelligence Research, a public speaker, and an active researcher.
His main research interests include machine learning, including swarm intelligence, deep learning, and adaptive learning in medicine, games, and chatbots. He has more than 100 peer reviews of scientific publications. He has supervised more than 110 student projects, including Master and Ph. D. theses within these topics, and more than 90 popular science public speaking events, mostly in Artificial Intelligence.
\end{IEEEbiography}








\end{document}